\documentclass[11pt]{article}

\usepackage[final]{acl}

\usepackage{times}
\usepackage{latexsym}
\usepackage[T1]{fontenc}
\usepackage[utf8]{inputenc}
\usepackage{microtype}
\usepackage{inconsolata}
\usepackage{graphicx}

\usepackage{booktabs}
\usepackage{multirow}
\usepackage{tabularx}
\usepackage{array}
\usepackage{amsmath}
\usepackage{amssymb}
\usepackage{enumitem}
\usepackage{xcolor}
\usepackage{makecell}
\usepackage[table]{xcolor}
\usepackage{todonotes}
\usepackage{subfig}
\usepackage[skins,breakable]{tcolorbox}
\usepackage{fontawesome5}
\usepackage{longtable}
\usepackage{booktabs}
\usepackage{tabularx}
\definecolor{barblue}{HTML}{4C78A8}
\definecolor{bargray}{HTML}{E6E6E6}
\definecolor{softgray}{HTML}{EBE0D2}
\definecolor{darkgreen}{RGB}{0,128,0}
\definecolor{darkred}{RGB}{180,0,0}

\usepackage{xspace}

\newcommand{\model}[1]{\textsc{#1}\xspace}

\newcommand{\gptfouro}{\model{GPT-4o}}

\newcommand{\gptfivefour}{\model{GPT-5.4}}

\newcommand{\gemma}{\model{Gemma-4-31B-it}}
\newcommand{\gptoss}{\model{GPT-OSS-20B}}
\newcommand{\llamaseventy}{\model{Llama-3.3-70B}}
\newcommand{\llamaeight}{\model{Llama-3.1-8B}}

\newcommand{\framework}[1]{\textsc{#1}\xspace}
\newcommand{\factscore}{\framework{FActScore}}
\newcommand{\medscore}{\framework{MedScore}}
\newcommand{\medfact}{\framework{MedFact}}

\newcommand{\dataset}[1]{\textsc{#1}\xspace}
\newcommand{\medsnip}{\dataset{MedSNIP}}            
\newcommand{\medsnipbench}{\dataset{MedSNIP-Bench}} 
\newcommand{\healthfc}{\dataset{HealthFC}}
\newcommand{\medhallu}{\dataset{MedHallu}}
\newcommand{\pubmedqa}{\dataset{PubMedQA}}

\newcommand{\medbullets}{\dataset{MedBullets}}
\newcommand{\kqa}{\dataset{K-QA}}
\newcommand{\usmle}{\dataset{USMLE}}

\newcommand{\mode}[1]{\textsc{#1}\xspace}
\newcommand{\claimonly}{\mode{Claim-Only}}
\newcommand{\fullcontext}{\mode{Full-Context}}

\newcommand{\aggrule}[1]{\textsc{#1}\xspace}
\newcommand{\orfalse}{\aggrule{Or-False}}
\newcommand{\andfalse}{\aggrule{And-False}}
\newcommand{\majorityrule}{\aggrule{Majority}}
\newcommand{\thresholdk}{\aggrule{Threshold-$k$}}

\newcommand{\fF}{$\text{F}_1^{F}$\xspace}

\title{MedSNIP: Building and Benchmarking Snippet-Level Granularity\\ for Medical Fact Verification}

\author{
\textbf{Hasan Iqbal}\textsuperscript{1} \quad
\textbf{Sarfraz Ahmad}\textsuperscript{1} \quad
\textbf{Hyunjae Kim}\textsuperscript{2} \quad
\textbf{Sihyeon Park}\textsuperscript{3} \\
\textbf{Junjie Liao\textsuperscript{4,5}} \quad
\textbf{Qingyu Chen\textsuperscript{2}} \quad
\textbf{Preslav Nakov\textsuperscript{1}} \quad
\textbf{Yuxia Wang\textsuperscript{5}} \\
\textsuperscript{1}Mohamed bin Zayed University of Artificial Intelligence,
\textsuperscript{2}Yale University,
\textsuperscript{3}Korea University \\
\textsuperscript{4}Beijing Normal University,
\textsuperscript{5}INSAIT, Sofia University, “St. Kliment Ohridski”\\
\parbox{\linewidth}{\centering
\texttt{hasan.iqbal@mbzuai.ac.ae}, \quad \texttt{yuxia.wang@insait.ai}  \\
[0.3em]
\faGlobe\ \href{https://mbzuai-nlp.github.io/MedSNIP/}{Project}
\quad
\faDatabase\ \href{https://huggingface.co/datasets/MBZUAI/MedSNIP}{\medsnip{}}
\quad
\faGithub\ \href{https://github.com/mbzuai-nlp/MedSNIP}{Code}
}
}

\begin{document}
\maketitle

\begin{abstract}
A medical claim's correctness often depends not on the claim alone, but on the clinical structure around it.
A claim may require a lab reference range, a causal or conditional link, or patient-specific details to be judged correctly, and atom-level decomposition can fragment these dependencies, leaving the verifier with clinically incomplete claims. We reformulate medical fact-checking around \emph{snippet-level} verification, where clause-grouped units preserve local clinical structure. We introduce \medsnipbench{}, a human-annotated benchmark for snippet-level medical fact verification, and \medsnip{}, an automatic snippet-generation pipeline. \medsnipbench{} covers 276 consumer-health and clinical-vignette responses, segmented into 2{,}524 snippets with dual in-general and in-patient-context labels and six structural pattern codes.
\medsnip{} is evaluated against human snippet boundaries on \medsnipbench{} and then used to generate snippet-level units for external corpora. Across \medsnipbench{}, \healthfc{}, and \medhallu{}, snippet-level verification preserves or improves false-class F1, with gains concentrated where answers are long enough to fragment and where the verifier is strong enough to exploit the recovered structure. The largest merge-pattern gain is on causal-conditional clinical chains. It also reduces verifier calls by 24--73\%, though the saving survives end-to-end only when decomposition is cheap, which an open-weight decomposer makes possible at no loss of chunking fidelity.
\end{abstract}
\section{Introduction}
\label{sec:intro}

Large language models (LLMs) are increasingly used to answer medical questions from patients, consumers, and clinicians~\citep{singhal2023large, singhal2025toward, manes-etal-2024-k, liu2025application, wang2024large}. A fluent but unsupported recommendation can influence patients' decisions about care, mislead downstream systems, and propagate through clinical reasoning~\citep{zhu2025can}.

Automated fact-checking is a safety requirement for medical LLMs, not an evaluation convenience. Currently, most fact-checking pipelines descend from \factscore{}~\citep{min2023factscore} and its successors.
They split a generation into minimal standalone \textit{atomic} claims, verify each claim independently, and aggregate the verdicts~\citep{chern2023factool, wei2024long, dhuliawala-etal-2024-chain, iqbal-etal-2024-openfactcheck}. This atom-level design is natural when each claim can be judged on its own. However, medical answers often violate this assumption. A lab interpretation may require both a measured value and a reference range. A diagnosis may require a pattern of findings rather than any finding in isolation. A recommendation may be true in general but wrong for the patient described in the question. In such cases, atomization can turn a clinically meaningful judgment into fragments that lack the information needed for verification. 

\begin{figure*}[t]
    \centering
    \includegraphics[width=0.96\linewidth]{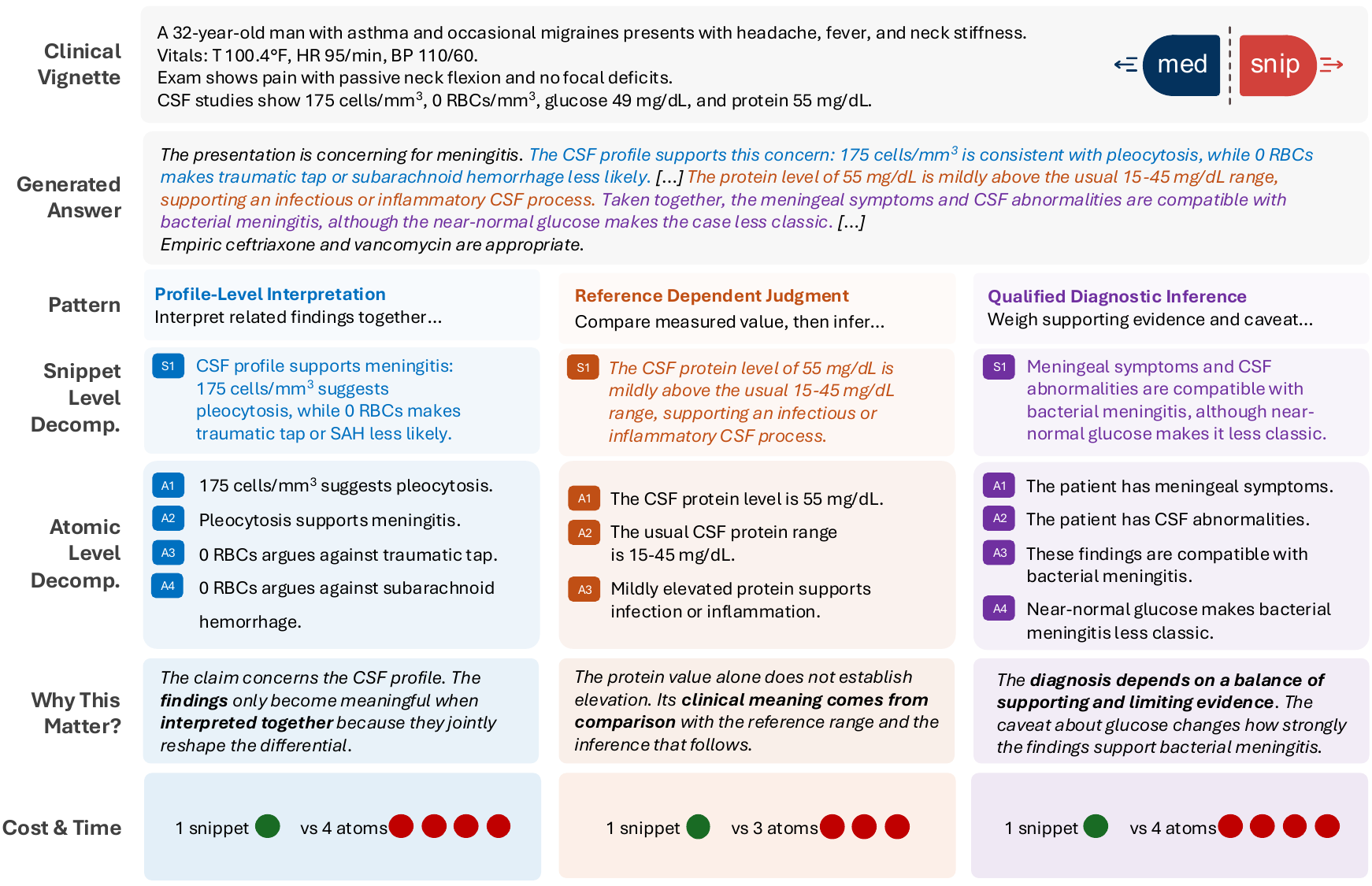}
    \caption{Motivating example from \medsnipbench{} showing how atomization can split clinically coherent judgments. Each column presents one judgment from a clinical-vignette answer. Snippet-level decomposition preserves the linked evidence, interpretation, and caveat as one verifiable unit, while atom-level decomposition fragments the same judgment into three or four separate claims.}
    \label{fig:motivation}
\end{figure*}

This motivates a coarser verification unit that preserves these dependencies. We call this unit a \emph{snippet}, a clinically meaningful group of related clauses that is smaller than a full answer but coarser than an atom. It groups the information needed to verify a clinical judgment, such as a finding and its interpretation, a lab value and its reference range, or a diagnosis and its caveat. Figure~\ref{fig:motivation} illustrates this on a clinical-vignette answer, where one judgment fragments into three or four atoms. This motivates treating granularity as a core design choice in medical fact-checking, not as pre-processing.

Motivated by these observations, we reformulate medical fact-checking around \emph{snippet-level verification}. Rather than decomposing responses into atomic claims or evaluating entire answers holistically, we group semantically and clinically related clauses into coherent snippets that preserve the context needed to verify clinical judgments. Each snippet is annotated for both general medical correctness and patient-specific correctness, capturing errors that may be medically plausible in general but unsafe for the particular patient context.

Existing medical fact-checking resources do not evaluate this formulation. \healthfc{} provides expert-labeled consumer-health claims and \medhallu{} provides a \pubmedqa{}-derived hallucination benchmark, but both evaluate near-atomic claims rather than clinically structured units~\citep{vladika-etal-2024-healthfc, pandit-etal-2025-medhallu}. Domain-tailored decomposers and verifier diagnostics similarly remain within the atom-level paradigm~\citep{huang2025medscore, he2025medfact}.
To fill this gap, we introduce \textbf{\medsnipbench{}}, a human-annotated benchmark for snippet-level medical fact-checking, together with \textbf{\medsnip{}}, an automatic snippet-generation pipeline. Together, this paper contributes:

\begin{itemize}[leftmargin=*,itemsep=2pt]
    \item A reformulation of medical fact-checking around snippet-level verification, where related clauses are grouped and checked as clinically coherent units that preserve relevant context.
    \item \medsnipbench{}, with 276 consumer-health and clinical-vignette responses, 2{,}524 human-annotated snippet boundaries, dual \textit{in-general} and \textit{in-patient-context} labels, and six structural pattern codes.
    \item \medsnip{}, an automatic snippet-generation pipeline evaluated against human snippets on \medsnipbench{} and used to generate snippet-level units for external corpora such as \healthfc{} and \medhallu{}.
    \item A cross-model, cross-dataset evaluation with paired bootstrap confidence intervals throughout, plus per-pattern analysis showing where snippet-level verification helps.
\end{itemize}
\section{Related Work}
\label{sec:related}

\paragraph{Atomic fact-checking:}
\factscore{} established a common paradigm for long-form factuality evaluation, where a generation is decomposed into atomic claims and each claim is checked against evidence~\citep{min2023factscore}.
Subsequent systems adopt claim-level verification pipelines~\citep{chern2023factool, wei2024long, gao-etal-2023-rarr, iqbal-etal-2024-openfactcheck, zerong-etal-2025-systematic, wang-etal-2025-openfactcheck, wang-etal-2024-factuality}.
Recent work questions whether the smallest claim is always the best verification unit.
\citet{wanner-etal-2024-closer} show that verifier accuracy depends on decomposition granularity, while \citet{lu-etal-2025-optimizing} learn a verifier-preferred atomicity policy that improves over default atomization.

Motivated by these findings, we take a simpler approach and show that a coarser, clause-grouped unit is sufficient, cheaper, and better aligned with medical reasoning.

\paragraph{Medical fact-checking:}
Medical factuality evaluation spans consumer-health and clinical question-answering settings.
\healthfc{} provides expert-labeled consumer-health claims with abstracts~\citep{vladika-etal-2024-healthfc}, while \medhallu{} extends \pubmedqa{} into a hallucination-detection benchmark with paired ground-truth and adversarially perturbed answers~\citep{pandit-etal-2025-medhallu}.
\medscore{} adapts \factscore{}-style decomposition to free-form medical answers and shows that a domain-tailored atomizer produces more valid claims than an atomizer~\citep{huang2025medscore}.
\medfact{} identifies an over-criticism failure mode in which stronger reasoning verifiers incorrectly mark correct medical text as false~\citep{he2025medfact}.
\citet{kim2025rethinking} show that retrieval covers only a minority of clinician-identified must-have information and that retrieval-augmented generation (RAG) can degrade factuality. \citet{gunjal-durrett-2024-molecular} make atomic claims interpretable through decontextualization, which is complementary to grouping rather than an alternative to it.

Rewriting ``\emph{A because B}'' as separate claims can miss errors when each part is plausible but their link is not. Existing methods improve claim-level medical fact-checking, but lack snippet structure and dual in-general and in-patient-context labels. We introduce \medsnipbench{}, which, to the best of our knowledge, is the first human-annotated snippet-level medical fact-checking dataset to preserve clinical claims and their context. Our results support snippet-level verification rather than relying only on atomized claims.
\section{\medsnipbench{}}
\label{sec:dataset}

\medsnipbench{} is built on the benchmark released by \citet{kim2025rethinking}, which pairs 100 consumer-health queries from \kqa{}~\citep{manes-etal-2024-k} with 100 \usmle-style clinical-vignette questions from \medbullets{}~\citep{chen-etal-2025-benchmarking} and carries expert true-or-false labels on atomic claims. We use 276 entries containing 5{,}755 expert-labeled atomic claims, partitioned into 140 consumer-health and 136 clinical-vignette entries using a word-length threshold on the original query. The parent corpus is bimodal in query length, so the partition is unambiguous (Appendix~\ref{app:data-subset}).

The atom-level binary factuality labels (i.e., true or false) are obtained from physician annotations inherited from \citet{kim2025rethinking}, while \medsnipbench{} then groups these atoms into clinically meaningful snippets and provides the corresponding snippet-level annotations.
Every snippet boundary, pattern code and correctness label in \medsnipbench{} was assigned manually.

\paragraph{Snippets and pattern taxonomy:}
The atomic-claim layer is often too fine-grained for medical content, because clinical judgments can depend on related clauses. To preserve these dependencies, we re-segment the corpus into \emph{snippets}, clause-grouped verifiable units designed to retain shared entities, clinical framing, and causal or conditional relationships. Analysis across the parent benchmark showed that snippet boundaries recur around six structural patterns. We identified these patterns through error analysis of an atomic-claim verification baseline, where recurring verification failures clustered into the same structural cases. Patterns A--C, illustrated in Figure~\ref{fig:motivation}, are merge patterns, where multiple atoms should be verified together as a single snippet, while Patterns D--F are keep-atomic patterns, where the atom remains the appropriate verification unit. The six structural patterns are summarized below:

\begin{itemize}[leftmargin=*,itemsep=1.5pt, topsep=3pt]
    \item \textbf{Pattern A} for enumerations of properties of a single subject (features, causes, or factors that only make sense as a set)
    \item \textbf{Pattern B} for causal or conditional chains whose links share evidence (clinical reasoning, including diagnostic logic, drug-effect chains, and threshold-dependent recommendations)
    \item \textbf{Pattern C} for conclusions together with the premises that warrant them (a diagnosis or recommendation whose verifiability depends on the supporting findings just stated).
    \item \textbf{Pattern D} for self-contained standalone facts or lab interpretations whose verifiability does not depend on surrounding clauses
    \item \textbf{Pattern E} for genuine topic shifts where merging would conflate distinct subjects
    \item \textbf{Pattern F} for distinct facts about the same subject that do not jointly support a single judgment, including isolated false atoms surrounded by true ones (a case where atomization protects the false claim from being absorbed into a true context).
\end{itemize}

\begin{table*}[!t]
\centering
\small
\renewcommand{\arraystretch}{1}
\resizebox{\textwidth}{!}{%
\begin{tabular}{@{}ll r rr rr c rrrrrr@{}}
\toprule
\multirow{2}{*}{\textbf{Dataset}}
& \multirow{2}{*}{\textbf{Subset / Split}}
& \multirow{2}{*}{\textbf{Entries}}
& \multicolumn{2}{c}{\textbf{Units}}
& \multirow{2}{*}{\textbf{\%F in-general}}
& \multirow{2}{*}{\textbf{\%F in-context}}
& \multirow{2}{*}{\textbf{Labels}}
& \multicolumn{6}{c}{\textbf{Patterns}} \\
\cmidrule(lr){4-5}
\cmidrule(lr){9-14}
&
&
& \textbf{Count}
& \textbf{Type}
&
&
&
& \textbf{A}
& \textbf{B}
& \textbf{C}
& \textbf{D}
& \textbf{E}
& \textbf{F} \\
\midrule
\rowcolor{softgray}
\multicolumn{14}{c}{\textbf{\medsnipbench{} Statistics}} \\
\midrule
\multirow{7}{*}{\textbf{\medsnipbench{}}}
& Parent corpus
& 276
& 5{,}755
& atom
& --
& --
& --
& --
& --
& --
& --
& --
& -- \\
& Consumer
& 140
& 1{,}091
& snippet
& 12.4
& 9.2
& dual
& 634
& 101
& 79
& 194
& 10
& 73 \\
& Vignette
& 136
& 1{,}433
& snippet
& 14.6
& 11.4
& dual
& 374
& 124
& 242
& 423
& 29
& 241 \\
\cmidrule(lr){2-14}
& Train
& 180
& 1{,}599
& snippet
& 13.6
& 10.5
& dual
& 640
& 143
& 209
& 385
& 26
& 196 \\
& Dev
& 51
& 494
& snippet
& 13.8
& 10.7
& dual
& 187
& 44
& 58
& 126
& 10
& 69 \\
& Test
& 45
& 431
& snippet
& 13.7
& 10.0
& dual
& 181
& 38
& 54
& 106
& 3
& 49 \\
\cmidrule(lr){2-14}
& \textbf{Total}
& \textbf{276}
& \textbf{2{,}524}
& \textbf{snippet}
& \textbf{13.6}
& \textbf{10.5}
& \textbf{dual}
& \textbf{1008}
& \textbf{225}
& \textbf{321}
& \textbf{617}
& \textbf{39}
& \textbf{314} \\
\midrule
\rowcolor{softgray}
\multicolumn{14}{c}{\textbf{External Datasets}} \\
\midrule
\multirow{3}{*}{\textbf{\healthfc{}}}
& Atoms
& 750
& 1{,}076
& atom
& --
& --
& --
& --
& --
& --
& --
& --
& -- \\
& Snippets (Mode 1)
& 750
& 821
& snippet
& --
& --
& single
& --
& --
& --
& --
& --
& -- \\
& Snippets (Mode 2)
& 750
& 761
& snippet
& --
& --
& single
& --
& --
& --
& --
& --
& -- \\
\cmidrule(lr){2-14}
\multirow{3}{*}{\textbf{\medhallu{}}}
& Atoms
& 2{,}000
& 3{,}798
& atom
& --
& --
& --
& --
& --
& --
& --
& --
& -- \\
& Snippets (Mode 1)
& 2{,}000
& 2{,}636
& snippet
& --
& --
& single
& --
& --
& --
& --
& --
& -- \\
& Snippets (Mode 2)
& 2{,}000
& 2{,}586
& snippet
& --
& --
& single
& --
& --
& --
& --
& --
& -- \\
\bottomrule
\end{tabular}%
}
\caption{\textbf{\medsnipbench{} and external dataset statistics.} The parent corpus is shown for reference. \textbf{\%F in-general} and \textbf{\%F in-context} are false-label rates, and A--F are structural pattern counts. For external datasets, both \medsnip{} modes and their shared atom baseline are reported (Section~\ref{subsec:medsnip-pipeline}). \healthfc{} counts follow question-to-proposition conversion. Dashes indicate unavailable or inapplicable fields.}
\label{tab:dataset-statistics}
\end{table*}

\paragraph{Dual correctness labels:} Each snippet receives two correctness judgments: \emph{(i) in-general} indicates whether the snippet is correct as a general medical statement. \emph{(ii) with-patient-context} indicates whether it is correct for the specific patient or query. This dual scheme captures cases that are generally true but contextually wrong, such as recommendations that are reasonable in general medicine but inappropriate for the vignette at hand. 

This distinction appears in practice in \medsnipbench{}, where 124 snippets, or 4.9\%, have divergent in-general and with-context labels. These cases are concentrated in clinical-vignette entries, consistent with contextual error being especially relevant in patient-specific settings. Full details and examples are in Appendix~\ref{app:guidelines}.

\subsection{Annotation}
\label{sec:dataset-annotation}

Six annotators with biomedical-NLP backgrounds segmented and labeled the data using a shared interface and written guidelines (see Appendix~\ref{app:guidelines}). Snippet boundaries and pattern codes were based on linguistic and logical judgments over the answer text, rather than clinical judgments requiring medical expertise.
Annotators also saw the parent corpus’s physician true-or-false labels for each atomic claim, anchoring the task in expert annotations. For each entry, an annotator first identified the shared context (Appendix~\ref{app:data-context}), such as patient demographics or query topic. They then applied the A–F pattern taxonomy to decide which claims should be merged into a snippet and which should remain separate. 

After grouping, the annotator revised each snippet into a self-contained statement and assigned two binary correctness labels. On a stratified inter-annotator agreement (IAA) set covering both source subsets, label outcomes, and merge patterns, agreement is substantial for the \textit{in-general} label and moderate for the harder \textit{with-patient-context} label, with Fleiss’ $\kappa$ of $0.662$ and $0.472$, respectively, following \citet{landis1977measurement}. Snippet boundary agreement is strong, with a mean pairwise adjusted Rand index (ARI) of $0.723$. (see Appendix~\ref{app:data-iaa}).

\begin{figure*}[!t]
\centering
\includegraphics[width=0.95\textwidth]{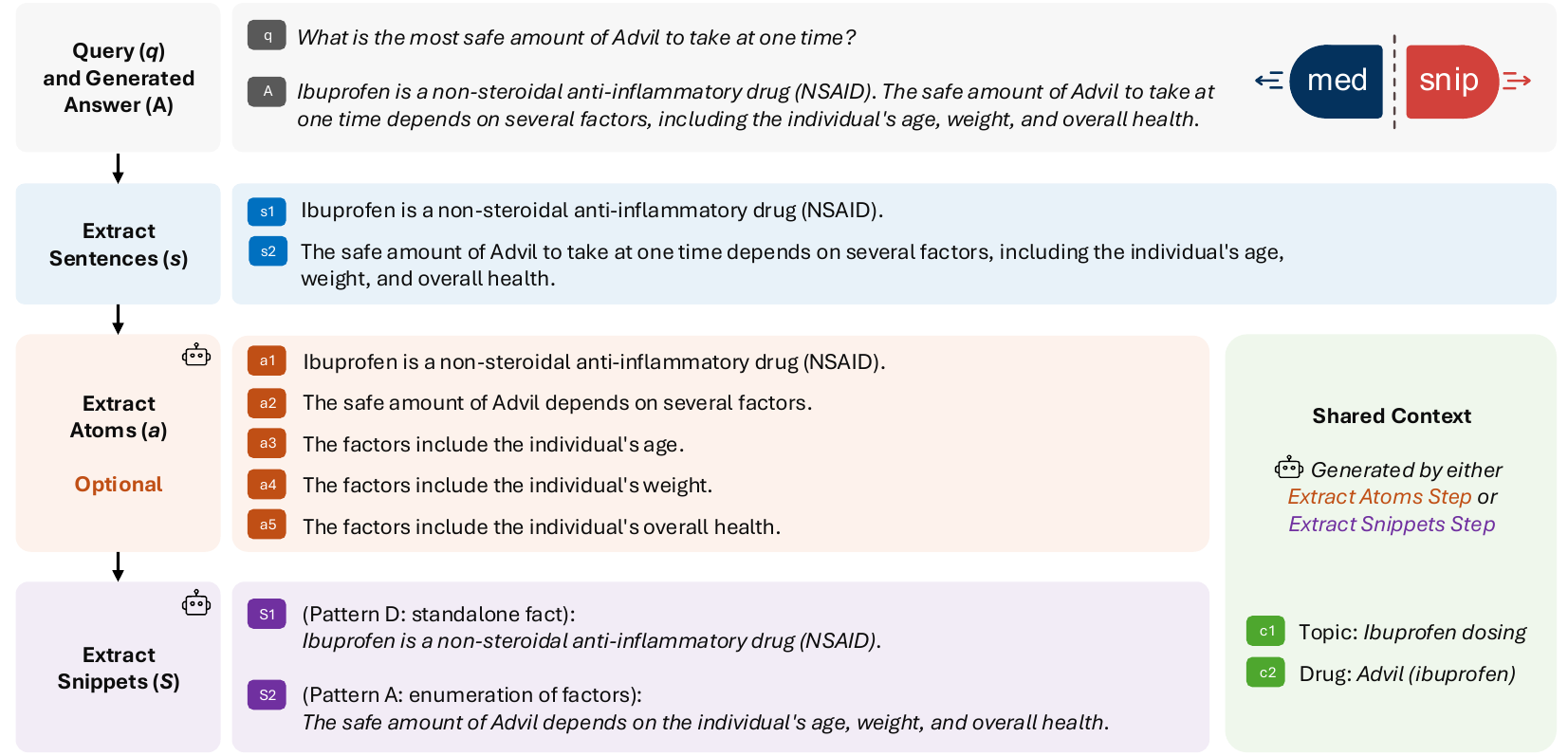}
\caption{\medsnip{} snippet-generation pipeline on a worked Advil example.
After deterministic sentence splitting, Mode~1 extracts atoms and shared context, then clusters atoms into snippets.
Mode~2 generates snippets directly from sentences.
Both modes return snippets with source pointers, A--F pattern codes, and shared context.}
\label{fig:medsnip-pipeline}
\end{figure*}

\subsection{Dataset Statistics}
\label{sec:dataset-stats}

After annotation, the parent corpus's 5{,}755 atomic claims map to 2{,}524 snippets. Of these, 1{,}554 are multi-atom and 970 are single-atom, with a median of 2 atoms per snippet. Because the dataset skews toward true snippets, we use \fF{}, defined as F1 on false claims, as the primary metric. The train, dev, and test partitions use entry-level splits that preserve source-subset proportions and snippet-level false rates across label dimensions (Appendix~\ref{app:data-split}). All 276 entries are used for the zero-shot snippet-vs-atom comparison. The retrieval-augmented verifier is developed on dev and confirmed on test. Consumer-health entries favor enumeration, while clinical vignettes contain more reasoning chains, diagnostic conclusions, and standalone interpretations. We use two external medical fact-checking datasets, \healthfc{}~\citep{vladika-etal-2024-healthfc} and \medhallu{}~\citep{pandit-etal-2025-medhallu}, to test whether the snippet-vs-atom comparison generalizes beyond \medsnipbench{}. 

From \healthfc{}, we use 327 claims with binary support labels and exclude 423 insufficient-evidence claims from F1 computation. From \medhallu{}, we use 2{,}000 binary items comprising 1{,}000 paired ground-truth and hallucinated answers derived from \pubmedqa{}~\citep{jin-etal-2019-pubmedqa}. Table~\ref{tab:dataset-statistics} summarizes corpus statistics, splits, structural patterns, and external datasets. Since the external datasets lack snippet boundaries, we apply \medsnip{} to generate snippets before evaluation. Atoms and snippets come from the same decomposition call, so both use identical input and differ only in verification unit. Because \healthfc{} presents claims as questions, we convert them to propositions before decomposition.

\section{Method}
\label{sec:method}

We compare medical fact-checking pipelines along three axes: \textit{(i)} verification unit, \textit{(ii)} verification context, and \textit{(iii)} verifier design. The main comparison isolates verification granularity by evaluating atoms and snippets. Given a medical answer $A$, a decomposer produces units $\{u_1,\ldots,u_n\}$, and a verifier labels each as \textsc{true} or \textsc{false}. The atom condition follows the \factscore{} lineage and verifies the smallest standalone factual statements, while the snippet condition verifies grouped clauses. For \medsnipbench{}, snippets are human-annotated. For \healthfc{} and \medhallu{}, they are generated using the \medsnip{} pipeline.

\subsection{\medsnip{} Pipeline}
\label{subsec:medsnip-pipeline}

\medsnip{} pipeline converts a long-form answer $A$ and optional query $q$ into snippets $S$, grouping the answer's own clauses under the question rather than any retrieved evidence. Each snippet includes source pointers, one of the A--F structural pattern codes, and shared context $ctx$.
Source pointers identify the atoms or sentences that support the snippet. Shared context is a short structured summary of the answer's topic, entities, and patient or query framing.
The LLM steps use this context to interpret otherwise underspecified units. As a first step, we apply deterministic sentence splitting after removing citation markers, bullets, numbering, and headers. Sentence indices give the decomposer fixed source pointers, which preserve provenance, let us check that every sentence is assigned, and provide the shared index space in which automatic and human snippets are compared. The splitter is rule-based rather than learned, so it introduces no additional model into the pipeline.

To compare automatic snippets with human snippets, we align source atoms to sentences using normalized-token overlap and substring matching, enabling sentence-set F1 evaluation without an additional model. The pipeline has two modes. Mode~1 (Atomize-then-group) first extracts atoms and then uses an LLM to cluster them into snippets, mirroring the benchmark construction process. Mode~2 (Snippet-direct) skips the atom intermediate and generates snippets directly from sentences.

Figure~\ref{fig:medsnip-pipeline} illustrates both modes using an Advil example, where listed factors are either extracted as atoms and grouped or grouped directly into snippets. Atomize-then-group uses two LLM calls and mirrors human annotation, while snippet-direct uses one. We use \gptfivefour{}~\citep{openai2026gpt54} as the decomposer across modes and datasets to avoid confounding the baselines. Section~\ref{subsec:exp-robustness} varies the decomposer across all six models. Full prompts appear in Appendix~\ref{app:prompts}. On the full \medsnipbench{}, atomize-then-group better matches human snippets than snippet-direct in sentence-level F1 ($0.756$ vs.\ $0.734$) and mean embedding similarity ($0.824$ vs.\ $0.794$). However, snippet-direct achieves higher \fF{} across all three datasets (Section~\ref{subsec:exp-robustness}). Boundary agreement and verification utility are therefore distinct, so we report both modes. Appendix~\ref{app:data-pipeline-quality} provides the full pipeline evaluation. Pattern codes are retained for analysis, while the main per-pattern results use human annotations from \medsnipbench{}.

\subsection{Verification Modes}
\label{subsec:verification-modes}

For each unit $u$, we evaluate two verification modes.
\emph{(i)}~{\claimonly{} verification} gives the verifier only $u$, testing whether the unit is self-contained.
\emph{(ii)}~{\fullcontext{} verification} gives the verifier $u$ with the original question and full generated answer.
This context is used only to disambiguate the unit, recover omitted entities, and resolve references; the verifier is instructed not to treat the generated answer as evidence.
This tests whether context lost during decomposition can be recovered at verification time.

\subsection{Retrieval-Augmented Verifier}
\label{subsec:verifier}

Single-call verifiers judge each unit from parametric knowledge alone.
This can be brittle for medical text, where verification may depend on drug indications, doses, reference ranges, guidelines, or evidence.
We therefore also evaluate a retrieval-augmented verifier.
It searches for evidence over multiple rounds, reasons over the accumulated evidence, and returns \textsc{true}, \textsc{false}, or \textsc{abstain} when the evidence is insufficient. The verifier maintains an evidence pool $\mathcal{E}_t$, initialized as:

\[
\mathcal{E}_0 = \emptyset
\]
At retrieval step $t$, the model writes a search query from the unit, original query, and current evidence:
\[
s_t = \mathrm{LLM}_{query}(u, q, \mathcal{E}_{t-1})
\]

Retrieval runs in parallel over Google Serper\footnote{\url{https://serper.dev/}} for general web evidence and the PubMed E-utilities API\footnote{\url{https://www.ncbi.nlm.nih.gov/books/NBK25501/}} for biomedical abstracts.
Each backend returns up to $k=3$ passages per query, and the evidence pool accumulates passages across rounds:

\[
\mathcal{E}_t =
\mathcal{E}_{t-1}
~\cup~\mathrm{Ret}_{web}(s_t)
~\cup~\mathrm{Ret}_{biomed}(s_t)
\]
The verifier reads the accumulated evidence and predicts a label with a confidence score $c_t \in [0,1]$ emitted with its verdict rather than by a separate calibrator:
\[
(\hat{y}_t, c_t) =
\mathrm{LLM}_{verify}(u, q, \mathcal{E}_t).
\]
Here, $\hat{y}_t \in \{\textsc{true}, \textsc{false}, \textsc{abstain}\}$.
Following \citet{xie-etal-2025-fire}, we use a confidence threshold $\tau$ and accept a \textsc{true} or \textsc{false} label only when $c_t \geq \tau$.
If $\hat{y}_t$ is \textsc{abstain} or $c_t < \tau$, the verifier treats the evidence as insufficient and performs another retrieval round, up to $T_{\max}=5$.
If no confident verdict is reached, a final prompt returns
\[
\hat{y}
=
\mathrm{LLM_{final}}(u, q, \mathcal{E}_{T_{\max}}),
\]
with the prompt biased toward \textsc{abstain} rather than an unsupported \textsc{true} or \textsc{false}.
In the abstention-dropped setting, units with \textsc{abstain} labels or confidence below $\tau$ are left unscored.
We also evaluate a calibrated-ensemble variant, where uncertain units fall back to the single-call baseline.
Section~\ref{subsec:exp-verifier} reports the resulting coverage-\fF{} trade-off. Full Prompts are provided in Appendix~\ref{app:prompts}.
\section{Experiments and Results}
\label{sec:experiments}

We evaluate whether snippet-level verification improves factuality across datasets, verifier models, and clinical structures.
Unless otherwise stated, experiments use \claimonly{} verification and report false-class F1, denoted \fF{}.

\begin{table*}[!t]
\centering
\small
\setlength{\tabcolsep}{4pt}
\begin{tabular}{l rrrr rrr rrr}
\toprule
& \multicolumn{4}{c}{\textbf{\medsnipbench{}}} & \multicolumn{3}{c}{\textbf{\healthfc{}}} & \multicolumn{3}{c}{\textbf{\medhallu{}}} \\
\cmidrule(lr){2-5}\cmidrule(lr){6-8}\cmidrule(lr){9-11}
\textbf{Model} & Atom & Snippet & Mode 1 & Mode 2 & Atom & Mode 1 & Mode 2 & Atom & Mode 1 & Mode 2 \\
\midrule
\gptfivefour{}-high & 0.321 & \cellcolor{green!15}\textbf{0.431} & \cellcolor{green!15}\textbf{0.429} & \cellcolor{green!15}\textbf{0.418} & 0.672 & 0.672 & \cellcolor{green!15}\textbf{0.693} & 0.694 & \cellcolor{green!15}\textbf{0.720} & \cellcolor{green!15}\textbf{0.721} \\
\gptfouro{}        & 0.344 & \cellcolor{green!15}\textbf{0.401} & \cellcolor{red!15}0.301 & \cellcolor{red!15}0.317 & 0.689 & \cellcolor{green!15}0.710 & \cellcolor{green!15}\textbf{0.736} & 0.559 & \cellcolor{green!15}0.568 & \cellcolor{green!15}\textbf{0.579} \\
\gemma{}     & 0.340 & \cellcolor{green!15}\textbf{0.389} & \cellcolor{green!15}0.350 & \cellcolor{green!15}0.377 & 0.664 & \cellcolor{green!15}0.682 & \cellcolor{green!15}0.690 & 0.612 & \cellcolor{green!15}\textbf{0.639} & \cellcolor{green!15}\textbf{0.648} \\
\gptoss{}        & 0.344 & \cellcolor{green!15}\textbf{0.401} & \cellcolor{red!15}0.330 & \cellcolor{green!15}0.366 & 0.680 & \cellcolor{red!15}0.678 & \cellcolor{green!15}0.689 & 0.623 & \cellcolor{green!15}0.627 & \cellcolor{green!15}0.636 \\
\llamaseventy{}      & 0.316 & \cellcolor{green!15}0.331 & \cellcolor{red!15}\textbf{0.235} & \cellcolor{red!15}0.280 & 0.678 & \cellcolor{green!15}0.700 & \cellcolor{green!15}0.693 & 0.507 & \cellcolor{red!15}\textbf{0.482} & \cellcolor{red!15}0.487 \\
\llamaeight{}       & 0.238 & \cellcolor{green!15}0.278 & \cellcolor{red!15}0.201 & \cellcolor{red!15}0.212 & 0.612 & \cellcolor{red!15}0.598 & \cellcolor{red!15}0.610 & 0.471 & \cellcolor{green!15}\textbf{0.533} & \cellcolor{green!15}\textbf{0.556} \\
\bottomrule
\end{tabular}
\caption{\fF{} for \claimonly{} verification across six models and three datasets. Green and red mark gains and losses over atoms, with significant differences in bold (95\% confidence interval). \medsnipbench{} uses human snippets, while both pipeline modes use drop-mixed projection. External units share a decomposition call.}
\label{tab:f1-cross-model}
\end{table*}

\subsection{Experimental Protocol}
\label{subsec:common-setup}

We evaluate on the three datasets defined in Table~\ref{tab:dataset-statistics}, \medsnipbench{}, \healthfc{}, and \medhallu{}.
On the external datasets, we use the atomize-then-group pipeline to produce snippets before verification. We evaluate six verifier models. The closed models are \gptfivefour{} with high reasoning effort and \gptfouro{} with temperature zero~\citep{openai2024gpt4ocard}.

The open-weight models are \gemma{}, \gptoss{}, \llamaseventy{}, and \llamaeight{}~\citep{google2026gemma4, openai2025gptoss120bgptoss20bmodel, grattafiori2024llama3herdmodels}. All models use the same prompts, with atom and snippet decompositions produced by \gptfivefour{} to isolate verification granularity. For \healthfc{} and \medhallu{}, \orfalse{} predicts an answer as \textsc{false} if any unit is predicted \textsc{false}. \medsnipbench{} requires no answer-level aggregation because labels are at the snippet level. Automatic snippets use drop-mixed projection from human atom labels. Significance uses 95\% bootstrap intervals over 10{,}000 paired entry- or claim-level resamples (see Appendix~\ref{app:experiments}).

\subsection{Snippet Verification Improves \fF{}}
\label{subsec:exp-f1}

Table~\ref{tab:f1-cross-model} compares atoms and snippets across six verifiers and three datasets. On \medsnipbench{}, human snippets outperform atoms for all verifiers, significantly for the four strongest. With automatic snippets, only \gptfivefour{}-high gains significantly, scoring $0.429$ versus $0.431$ on human snippets. Four other verifiers fall below their atom baselines, significantly for \llamaseventy{}. Thus, the human-automatic gap is verifier-dependent. Snippet-direct outperforms atomize-then-group for all but the strongest verifier and on the external datasets. It yields two significant \healthfc{} gains and four of eight significant \medhallu{} gains, while atomize-then-group yields none on \healthfc{}. Gains grow with verifier strength and source structure (Section~\ref{subsec:exp-structure}). Snippets also reduce verifier calls. On \medsnipbench{}, human snippets reduce units by 56\%, while atomize-then-group and snippet-direct reduce them by 73\% and 63\%. The two modes reduce units by 24\% and 29\% on \healthfc{}, and 31\% and 32\% on \medhallu{}. Since snippets are longer, \medsnipbench{} verification cost falls by 54\%, versus 73\% for units. Appendix~\ref{app:exp-cost} reports end-to-end costs.

\paragraph{Result analysis:} The cross-model results support the main granularity claim, with a boundary condition. Snippet-level verification reduces calls by grouping units and improves \fF{} across verifier families, but the improvement is not uniform. It is largest where the source answer is long enough for atomization to fragment a clinical judgement and where the verifier can reason over the recovered structure. Where either condition is absent the effect narrows and can reverse, significantly so for \llamaseventy{} on both \medsnipbench{} and \medhallu{}. At the same time, the automatic-snippet columns show that segmentation quality matters. The strongest verifier is less affected by pipeline-generated snippets, while weaker verifiers show larger drops from human-snippet performance. Thus, snippet-level verification is a granularity choice whose benefit depends on both the verification unit and the verifier's ability to reason over it. The advantage is not unconditional. Under \fullcontext{} verification with weaker models, atoms can match or beat snippets, see Appendix~\ref{app:experiments} for more details.

\subsection{Structure and Snippet Gains}
\label{subsec:exp-structure}

Snippet-level verification can only help when there is meaningful structure for atomization to fragment. The three corpora differ substantially in this respect, providing a natural explanation for the variation in gains across datasets.

\begin{table}[t]
\centering
\small
\setlength{\tabcolsep}{4pt}
\begin{tabular}{lrrrr}
\toprule
\textbf{Corpus} & \textbf{Words} & \textbf{Sent.} & \textbf{Merge} & \textbf{$\Delta$ \fF} \\
\midrule
\medsnipbench{} & 244 & 11 & 3.37 & \cellcolor{green!15}$+0.114$ \\
\medhallu{}     &  29 &  1 & 1.44 & \cellcolor{green!10}$+0.026$ \\
\healthfc{}     &  11 &  1 & 1.31 & $+0.000$ \\
\bottomrule
\end{tabular}
\caption{Source length and grouping opportunity per corpus, with the \gptfivefour{} snippet--atom gap. All three columns use atomize-then-group, so the comparison is matched.}
\label{tab:corpus-structure}
\end{table}

Table~\ref{tab:corpus-structure} relates source length and merge ratio to the snippet-to-atom gap under \gptfivefour{}. \medsnipbench{} has the most structure, with medians of 244 words, 11 sentences, and 3.37 atoms per snippet, and the largest gain of $+0.114$ \fF{}. In contrast, 98\% of \healthfc{} claims are single sentences, yielding little structure and no gain. \medhallu{} lies between both datasets in structure and improvement. This ordering holds only with a strong verifier. Weaker open-weight verifiers can lose accuracy on richer \medsnipbench{} snippets (Table~\ref{tab:f1-cross-model}). Snippet-level verification therefore requires useful source structure and a verifier capable of using it. With three corpora, we report this ordering without fitting a formal relationship.

\subsection{Per-Pattern Analysis}
\label{subsec:exp-patterns}

The \medsnipbench{} annotations show where snippets help. Table~\ref{tab:patterns} reports pattern-level \fF{} using \gptfivefour{} with \claimonly{}. Snippets outperform atoms on all patterns except the small topic-shift category E. The largest merge-pattern gain is for Pattern B, causal-conditional chains, at $\Delta=+0.083$, suggesting that atomization harms verification of clause relations. Pattern E favors atoms, but contains only 39 snippets and its confidence interval crosses zero. For single-atom patterns D and F, boundaries often coincide, so gains may reflect cleaner snippet text rather than granularity. Patterns A, B, and C provide a clearer test. Results also depend on the setting. With \fullcontext{} or weaker verifiers, several gaps narrow or reverse, including Pattern B under \fullcontext{} with \gptfouro{} (Appendix~\ref{app:experiments}).

\begin{table}[!t]
\centering
\small
\setlength{\tabcolsep}{5pt}
\begin{tabular}{l r rr r}
\toprule
\textbf{Pattern} & \textbf{$n$} & \textbf{Snip} & \textbf{Atom} & \textbf{$\Delta$} \\
\midrule
A enumeration   & 1,008 & 0.353 & 0.293 & \cellcolor{green!15}\textbf{+0.059} \\
B causal/cond.  &  225 & 0.387 & 0.304 & \cellcolor{green!15}+0.083 \\
C concl.+prem.  &  321 & 0.442 & 0.391 & \cellcolor{green!10}+0.051 \\
D standalone    &  617 & 0.384 & 0.303 & \cellcolor{green!15}\textbf{+0.081} \\
E topic-shift   &   39 & 0.143 & 0.211 & \cellcolor{red!8}$-$0.068 \\
F distinct fact &  314 & 0.642 & 0.586 & \cellcolor{green!10}+0.056 \\
\midrule
Overall         & 2524 & 0.427 & 0.359 & \cellcolor{green!15}\textbf{+0.068} \\
\bottomrule
\end{tabular}
\caption{Per-pattern \fF{} on \medsnipbench{} using \gptfivefour{} under \claimonly{} verification.
Bold $\Delta$ marks significance under 95\% bootstrap confidence intervals.}
\label{tab:patterns}
\end{table}

\begin{table}[!t]
\centering
\small
\setlength{\tabcolsep}{4pt}
\begin{tabular}{lrrrr}
\toprule
\textbf{Decomposer} & \textbf{Sent F1} & \textbf{Merge} & \textbf{$\Delta$ \fF} & \textbf{Cost} \\
\midrule
\gptfivefour{} (high) & \textbf{0.756} & 3.37 & \cellcolor{green!15}$+0.099$ & \$23.63 \\
\gemma{}      & \textbf{0.756} & 2.08 & \cellcolor{green!15}$+0.061$ & \$0.29 \\
\llamaseventy{}       & 0.742 & 1.98 & \cellcolor{green!15}$+0.095$ & \$0.21 \\
\gptfouro{}         & 0.722 & 2.31 & \cellcolor{green!15}$+0.092$ & \$5.79 \\
\gptoss{}         & 0.706 & 2.59 & \cellcolor{green!15}$+0.055$ & \$0.20 \\
\llamaeight{}        & 0.695 & 1.37 & \cellcolor{green!15}$+0.045$ & \$0.03 \\
\bottomrule
\end{tabular}
\caption{Decomposer quality and downstream effect on \medsnipbench{}, with \gptfivefour{} verifying.
Every gap is measured against the same expert-atom baseline and every one excludes zero.}
\label{tab:decomposer-robustness}
\end{table}

\paragraph{Result analysis:} The results clarify why snippets help. The largest merge-pattern gain occurs on causal-conditional chains, where a conclusion depends on premises, conditions, or mechanisms.

These are the cases where atomization can separate the statement being checked from the information that makes it verifiable. Enumeration and conclusion-with-premises patterns also favor snippets, though with smaller gains.
The positive deltas for standalone and distinct-fact patterns should be interpreted more cautiously, since those gains may reflect cleaner snippet wording rather than grouping itself. Overall, the pattern analysis supports the central mechanism that snippets help most when the clinical judgment is distributed across related clauses.

\subsection{Robustness of the Decomposer}
\label{subsec:exp-robustness}

To test dependence on \gptfivefour{}, we generate snippets with all six models and verify them with all six verifiers in both pipeline modes. Table~\ref{tab:decomposer-robustness} reports the \gptfivefour{} verifier column, while Appendix~\ref{app:matrix} gives all 72 combinations. The table includes boundary fidelity, merge ratio, the snippet-direct gap against expert atoms, and decomposition cost for 276 entries. Boundary fidelity does not follow model scale. On the 275 entries completed by both models, \gemma{} matches \gptfivefour{} at $0.756$, with higher precision ($0.778$ vs.\ $0.723$), while costing \$0.29 instead of \$23.63. \llamaseventy{} also outranks \gptfouro{}. Rankings instead follow merge ratio. \gptfivefour{} groups most at 3.37 atoms per snippet, increasing recall to $0.898$ but lowering precision. \llamaeight{} groups least at 1.37 and reverses this balance. With \gptfivefour{} verification, every decomposer outperforms expert atoms, with all gaps significant under an entry-clustered bootstrap. \llamaeight{} gains $+0.045$ versus \gptfivefour{}'s $+0.099$, so decomposer variation is smaller than the gain over atoms. Boundary fidelity and downstream utility differ. \gemma{} ties for the best boundary agreement but has one of the smallest verification gains, echoing Section~\ref{subsec:medsnip-pipeline}. 

With only six decomposers, we report both rankings without fitting a correlation. Across the full matrix, verifier capability matters more than decomposer identity. Averaged over both modes, verifier column means span $0.113$ \fF{}, while decomposer row means span $0.019$, a sixfold difference. This favors deployment because decomposition runs once per answer, while verification runs once per unit.

\paragraph{Wording against grouping:}
Patterns D and F contain one atom each, so any gain there cannot come from grouping.
We isolate the two effects by rewriting each atom with the pipeline's decontextualization rules and no grouping, then verifying the rewritten atom.
Decontextualization alone recovers 94\% of the gain on the keep-atomic patterns and 26\% on the merge patterns, leaving 74\% of the merge-pattern gain attributable to grouping.
Wording explains least on Pattern A at 21\%, then Pattern B, the causal-conditional chains that motivate the method, at 25\%.
The full three-way comparison is in Appendix~\ref{app:normalisation}.

\paragraph{Cost with decomposition included:}
Reporting verifier calls alone understates what snippet-level verification costs, since building snippets takes an extra call per answer in atomize-then-group. Charging each pipeline for its own decomposition reverses the headline for the configuration used above. With \gptfivefour{} decomposing, snippet verification costs 19.1\% more end to end than atom verification, because the second decomposition call outweighs the verifier calls it saves. With \gptoss{} decomposing, the same comparison saves 34.1\%. The open-weight result above is therefore not only a robustness check. It is what makes the cost reduction survive honest accounting. Appendix~\ref{app:exp-cost} gives the full breakdown.

\subsection{Retrieval-Augmented Verification Provides a Coverage Lever}
\label{subsec:exp-verifier}

Table~\ref{tab:verifier} compares the retrieval-augmented verifier with the single-call baseline on \medsnipbench{} dev and test. At near-full coverage, retrieval matches but does not improve over the baseline. Higher confidence thresholds trade coverage for \fF{}. At $\tau=0.90$, \fF{} reaches $0.609$ on dev at 63\% coverage and $0.552$ on test at 61\%. The robust finding is therefore a coverage-\fF{} trade-off, not a full-coverage improvement. Full-coverage ensembles and subset routers improved dev results but did not replicate on test and are reported as negative results in Appendix~\ref{app:verifier}.

\begin{table}[!t]
\centering
\small
\setlength{\tabcolsep}{4pt}
\begin{tabular}{lrrrr}
\toprule
\textbf{Metric} & \textbf{Baseline} & \textbf{Verifier, full} & \textbf{$\tau \geq 0.85$} & \textbf{$\tau \geq 0.90$} \\
\midrule
Dev \fF{}  & 0.518 & 0.518 & \cellcolor{green!10}0.562 & \cellcolor{green!15}\textbf{0.609} \\
Dev cov.   & \cellcolor{gray!10}100\% & 96\%  & 83\%  & 63\% \\
\midrule
Test \fF{} & 0.491 & 0.497 & \cellcolor{green!10}0.504 & \cellcolor{green!15}\textbf{0.552} \\
Test cov.  & \cellcolor{gray!10}100\% & 97\%  & 83\%  & 61\% \\
\bottomrule
\end{tabular}
\caption{Coverage-\fF{} trade-off on \medsnipbench{} with \gptfivefour{}. Thresholded results exclude abstained and low-confidence units.}
\label{tab:verifier}
\end{table}

\paragraph{Results analysis:}
Retrieval augmentation serves a different role from snippets. It does not improve on the single-call baseline at full coverage, but enables confidence-based abstention. Excluding uncertain cases raises \fF{} on dev and test while reducing coverage, which may help when evidence and abstention matter. The fact that the full-coverage variants did not replicate also argues against treating retrieval augmentation as the source of the paper's main improvement.

\section{Conclusion and Future Work}
\label{sec:conclusion}

Medical fact-checking depends not only on the verifier, but also on the unit being verified.
We introduced \medsnipbench{}, a human-annotated benchmark for snippet-level medical fact-checking, and \medsnip{}, an automatic pipeline for generating snippet-level verification units.
Across \medsnipbench{}, \healthfc{}, and \medhallu{}, snippet-level verification improves false-class F1 under capable verifiers while reducing verifier calls.
The strongest and most interpretable gains occur when atomization separates causal, conditional, or premise-supported clinical reasoning into isolated claims.
This benefit depends on available structure and the verifier’s ability to use it.
These results suggest that verification granularity should be treated as a core design choice in medical fact-checking pipelines, not a fixed preprocessing default.

In future work, we plan to explore adaptive granularity policies that choose between atomic, snippet-level, and document-level verification based on clinical reasoning structure and verifier uncertainty, rather than relying on a fixed decomposition strategy. We further plan to extend \medsnip{} beyond English and integrate richer clinical evidence sources, such as guidelines, electronic health record–style data, and longitudinal patient context, to better evaluate factuality in realistic clinical decision-support settings.
\section*{Limitations}
\label{sec:limitations}

\paragraph{Scope:} Our experiments focus on English medical fact-checking datasets. Other languages and domains may require different granularity choices, especially when claims are more independent or when retrieval coverage differs substantially. We view snippet-level verification as a design choice for medically structured claims, not as a universal replacement for atom-level verification.

\paragraph{Model and aggregation dependence:} The main experiments cover six verifier models, with \gptfivefour{} used for decomposition across datasets. Absolute \fF{} scores depend on verifier strength, and the relative atom--snippet gap can vary with prompt context and aggregation rule. In particular, weak verifiers in \fullcontext{} mode can reduce or reverse the snippet advantage. We report aggregation sensitivity and verifier ablations in Appendix~\ref{app:experiments}.

\paragraph{Granularity and text quality:} When snippet and atom boundaries coincide, gains may come from clearer wording, such as resolved references, rather than granularity. Evidence for granularity is strongest for merge patterns, including enumerations, causal or conditional chains, and conclusions with premises. Appendix~\ref{app:normalisation} separates these effects by rewriting atoms without grouping. Wording explains nearly all of the keep-atomic gain but only about a quarter of the merge gain. Because repeated verification of identical text changes \fF{} by $0.013$, smaller differences may reflect variation rather than an effect.

\paragraph{Sample size:} \medsnipbench{} contains 276 entries and 2{,}524 snippets, which is enough to separate the largest pattern categories but not the smallest. Pattern E holds 39 snippets, and its negative delta has a confidence interval spanning $[-0.300, +0.095]$, so we draw no conclusion from it. The external evidence is similarly bounded, since \healthfc{} scores only the 327 claims that carry binary support labels.
Section~\ref{subsec:exp-structure} reports an ordering across three corpora rather than a fitted relationship for the same reason.

\paragraph{Pipeline and retrieval design:} External-dataset snippets are generated automatically with a single LLM-based decomposer. Section~\ref{subsec:exp-robustness} varies the decomposer over all six models, but only on \medsnipbench{}, so the decomposer sweep and the external corpora do not overlap.

Similarly, our retrieval-augmented verifier is intentionally simple and uses general web and biomedical-literature retrieval rather than specialized clinical-evidence sources. Stronger retrieval may improve coverage and absolute \fF{}, but retrieval design is not the main focus of this work.

\paragraph{Development and test use:}
We use the held-out test split to confirm patterns observed during development.
Two development-time variants, a calibrated ensemble and a subset-conditional router, did not replicate on test.
We report these negative results in Appendix~\ref{app:verifier}.

\section*{Ethics and Broader Impact}
\label{sec:ethics}

This work studies automated factuality evaluation for medical text generation. It is intended to support auditing and research on medical LLMs, not to provide medical advice, diagnosis, treatment recommendations, or autonomous clinical decision support. The systems evaluated here should be used only with appropriate expert oversight. A central motivation is harm reduction. Unsupported medical recommendations can mislead patients, clinicians, and downstream systems, and more reliable fact-checking may help identify such failures earlier. At the same time, automated verification is imperfect. A verifier may accept false claims, reject correct claims, or abstain on clinically important cases. Our results should therefore be interpreted as progress on evaluation methodology, not as evidence that automated medical fact-checking is solved.

\medsnip{} is derived from previously released medical QA resources and does not contain real patient records, protected health information, or identifiable clinical documentation. The clinical-vignette examples are synthetic educational cases rather than electronic health records. Nevertheless, the dataset may reflect biases in medical education materials, English-language evidence sources, and annotation practices. The annotation task also involves judgment calls about snippet boundaries and contextual correctness. We provide guidelines and report inter-annotator agreement, but other annotator groups or healthcare settings may choose different decompositions. We release the annotation guidelines, data, and code for inspection and reuse. Efficient verification could also scale low-quality medical content or overconfident evaluation. Snippet-level verification should support auditing, not replace expert review.

\bibliography{custom}

\newpage

\appendix
\clearpage
\section{Data Construction Details}
\label{app:data}

This appendix documents the construction of \medsnipbench{}, including how we partitioned the parent corpus, created the train, dev, and test splits, measured inter-annotator agreement, and evaluated how faithfully the \medsnip{} pipeline reproduces human snippet structure.

\subsection{Subset Selection}
\label{app:data-subset}

\begin{figure}[t]
\centering
\subfloat[Query word-length distribution.]{%
  \includegraphics[width=\columnwidth]{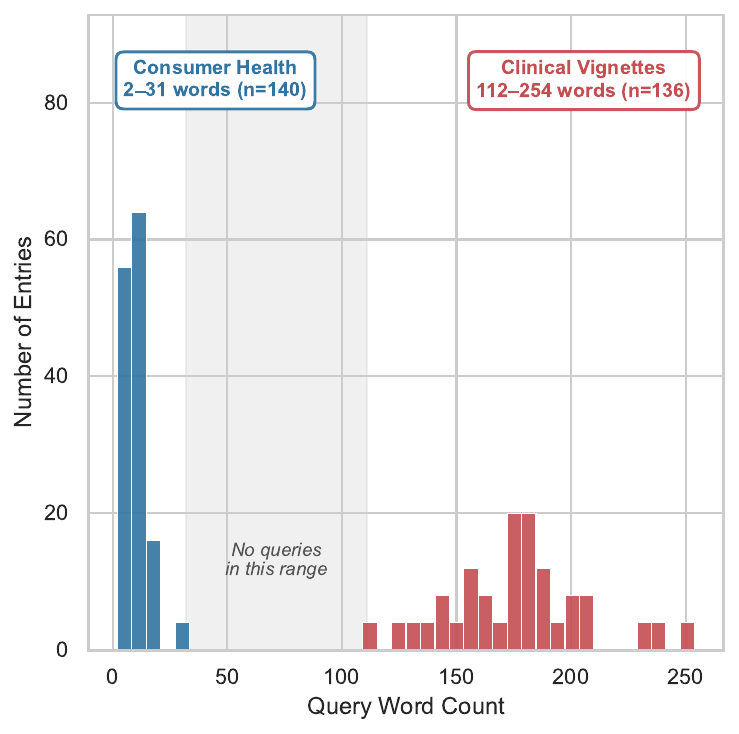}%
  \label{fig:subset-reasoning-a}}\\
\subfloat[Atomic-claim composition by subset.]{%
  \includegraphics[width=\columnwidth]{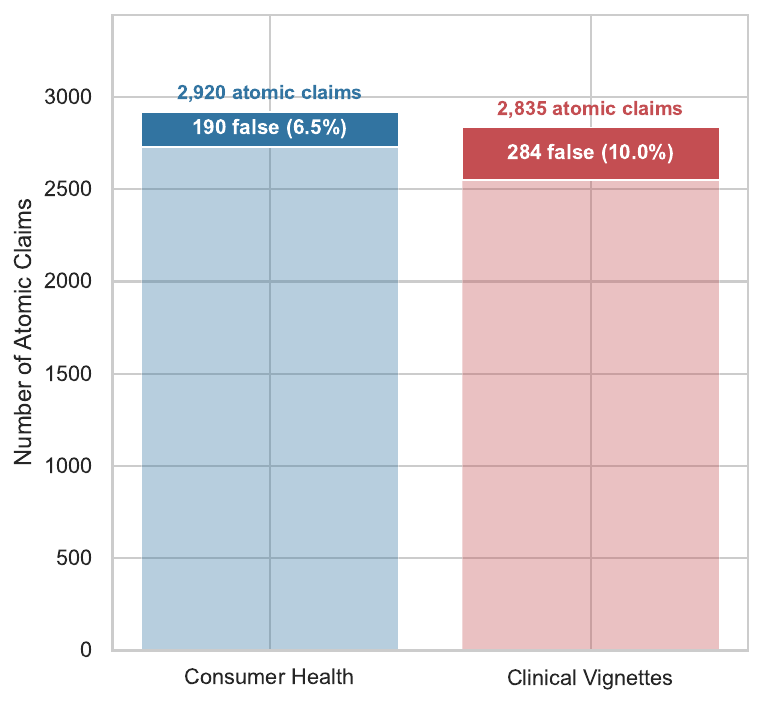}%
  \label{fig:subset-reasoning-b}}
\caption{Subset selection from the parent corpus.}
\label{fig:subset-reasoning}
\end{figure}

The \citet{kim2025rethinking} corpus blends short consumer-health questions from \kqa{}\citep{manes-etal-2024-k} with longer \usmle{}-style clinical-vignette questions from \medbullets{}\citep{chen-etal-2025-benchmarking}. Query length is bimodal, with consumer-health queries ranging from 2 to 31 words and clinical vignettes from 112 to 254 words, leaving no queries in the 32–111-word gap (Figure~\ref{fig:subset-reasoning}a). We use a threshold of 80 words to split the 276 entries into 140 consumer-health and 136 clinical-vignette entries. Any threshold between 32 and 111 yields the same partition.

The two subsets differ in verification demands as well as length. Consumer-health queries are short questions about a drug, dose, interaction, or symptom, with answers typically grounded in standard-of-care knowledge. Clinical vignettes provide a structured patient history and ask for a diagnosis, next test, or treatment, so the answer connects premises whose correctness depends on the patient’s specific context rather than general medical facts alone.

\paragraph{Representative queries:}
A typical consumer-health query is a single short sentence:
\begin{quote}\itshape\small
``Can I take Nyquil and Benadryl at the same time?''
\end{quote}
A typical clinical-vignette query is a paragraph-length scenario followed by a question:
\begin{quote}\itshape\small
``A 1-year-old girl is brought to a neurologist due to increasing seizure frequency over the past 2 months. She recently underwent a neurology evaluation which revealed hypsarrhythmia on EEG with multifocal spikes\ldots{} Her medications consist of lamotrigine and valproic acid\ldots{} What is the most appropriate next step in management?''
\end{quote}
The two query styles drive snippet patterns: consumer queries skew toward Pattern A (enumeration of features or factors), while vignettes skew toward Patterns B and C (causal--conditional chains and conclusion-with-premises). The pattern distribution in Table~\ref{tab:dataset-statistics} reflects this asymmetry.

Vignette responses carry a higher atom-level false-rate (10.0\%) than consumer responses (6.5\%), consistent with the harder clinical-reasoning content (Figure~\ref{fig:subset-reasoning}b). All downstream subset-conditional analyses preserve this distinction.

\subsection{Train / Dev / Test Split}
\label{app:data-split}

We stratify the 276 entries by source subset and a coarse in-general false-rate bucket, then sweep random seeds to minimize snippet-level false-rate differences across the three label dimensions while maintaining balanced splits. Seed 202 provides the best balance, with a worst-case spread of only $1.1$ percentage points across splits. Entry and snippet counts for each split are reported in Table~\ref{tab:dataset-statistics}.

\clearpage

\begin{figure}[t]
\centering
\subfloat[Per-subset entry counts across splits.]{%
  \includegraphics[width=0.86\columnwidth]{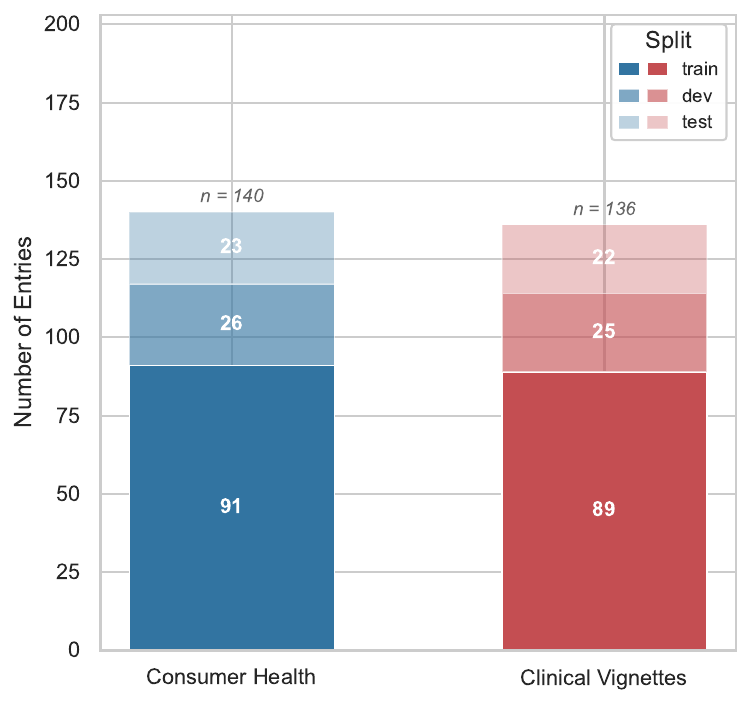}%
  \label{fig:split-reasoning-a}}\\
\subfloat[Snippet-level false-rate per (subset $\times$ split).]{%
  \includegraphics[width=0.86\columnwidth]{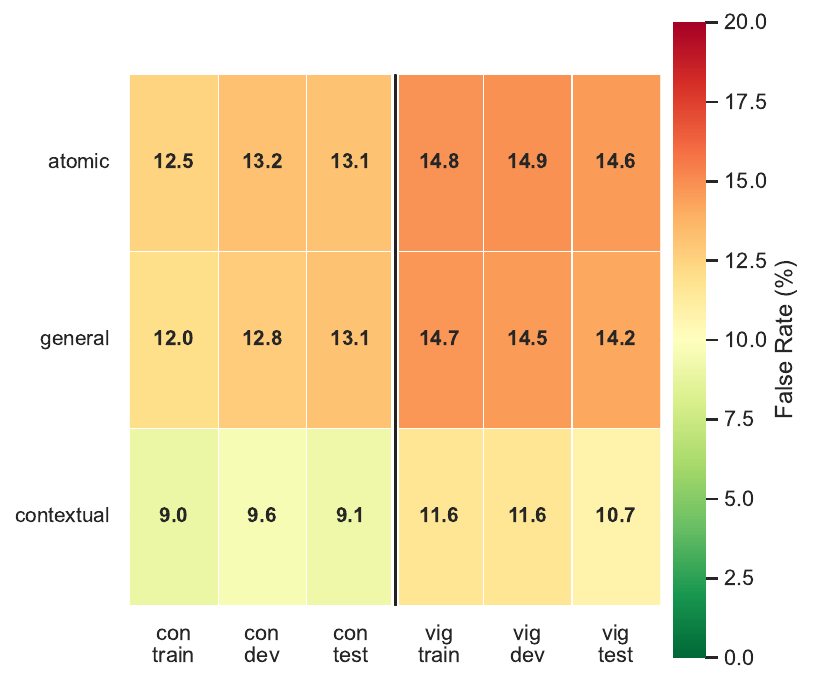}%
  \label{fig:split-reasoning-b}}
\caption{Train / dev / test split.}
\label{fig:split-reasoning}
\end{figure}

The resulting partition preserves the original consumer-to-vignette ratio across splits (Figure~\ref{fig:split-reasoning}a). Snippet-level false-rates are tightly matched across splits within each subset on all three label dimensions (Figure~\ref{fig:split-reasoning}b).

\subsection{Shared Context}
\label{app:data-context}

\begin{figure*}[t]
\centering
\subfloat[Grouping agreement (ARI per entry).]{%
  \includegraphics[width=0.32\textwidth]{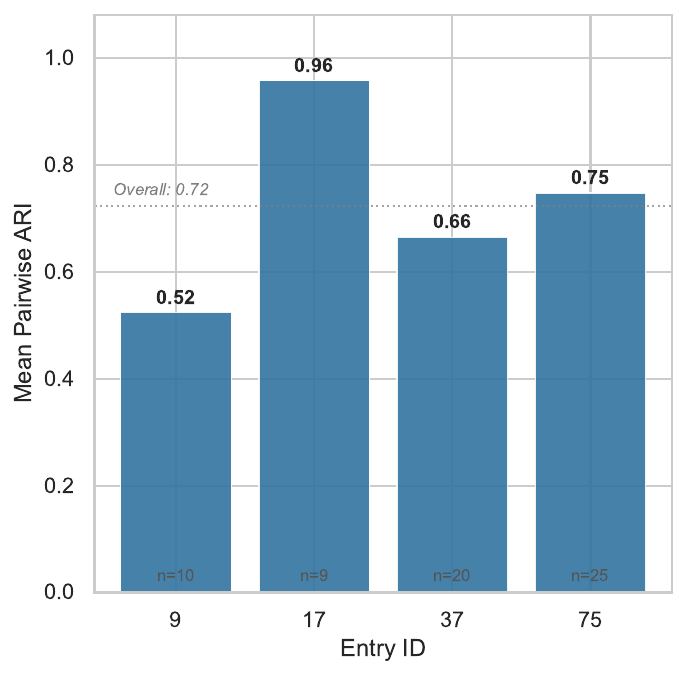}%
  \label{fig:iaa-reasoning-a}}
\hfill
\subfloat[Fleiss' $\kappa$ per labeled field.]{%
  \includegraphics[width=0.32\textwidth]{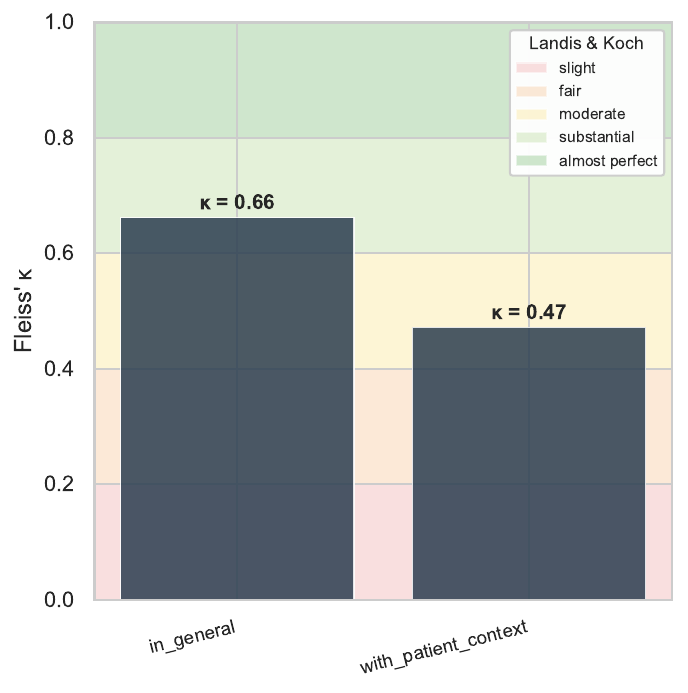}%
  \label{fig:iaa-reasoning-b}}
\hfill
\subfloat[Pairwise Cohen's $\kappa$ for in-general.]{%
  \includegraphics[width=0.32\textwidth]{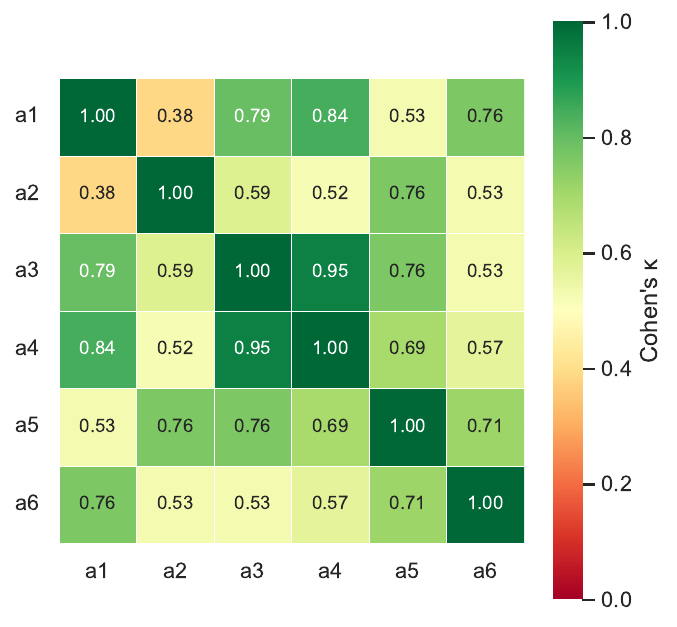}%
  \label{fig:iaa-reasoning-c}}
\caption{Inter-annotator agreement on the 64-snippet IAA batch annotated independently by all six annotators.}
\label{fig:iaa-reasoning}
\end{figure*}

Each entry includes a \textit{shared context} block, a structured key–value summary of the query and response framing that applies to all snippets.
It anchors decontextualized snippets in the clinical or topical scenario, supporting verification against the relevant patient or query. The schema is subset-dependent. Clinical-vignette entries use keys such as \texttt{age}, \texttt{sex}, \texttt{chief\_complaint}, \texttt{medical\_history}, \texttt{medications}, \texttt{vitals}, \texttt{treatment\_history}, and \texttt{correct\_answer}. Consumer-health entries use lighter keys such as \texttt{topic} and \texttt{drug\_a}/\texttt{drug\_b}.The schema is open, allowing annotators to add keys when downstream snippets refer to additional information.

\medsnip{} pipeline preserves this convention. \mbox{Mode 1} and \mbox{Mode 2} both emit a shared-context block alongside the snippet list (Section~\ref{subsec:medsnip-pipeline}), so automatic and human-annotated entries are interchangeable downstream. Verification prompts may reference the shared context to recover entities that are absent from the snippet text, which matters most for the with-patient-context label where the relevant frame is not in the snippet itself.

\subsection{Inter-Annotator Agreement}
\label{app:data-iaa}

Inter-annotator reliability for \medsnipbench{} was computed on a 64-snippet IAA batch annotated independently by all six annotators. Figure~\ref{fig:iaa-reasoning} shows chunking agreement per entry, Fleiss’ $\kappa$ for each labeled field, and pairwise Cohen’s $\kappa$ for the in-general label.

\paragraph{Binary in-general label:}
Fleiss’ $\kappa = 0.662$ across six annotators (Figure~\ref{fig:iaa-reasoning-b}). Pairwise Cohen’s $\kappa$ ranged from $0.385$ to $0.947$ across 15 annotator pairs, with a mean of $0.663$ (Figure~\ref{fig:iaa-reasoning-c}). All six annotators agreed on 52 of 64 items ($81.3\%$).

\paragraph{Snippet chunking:}
Adjusted Rand index (ARI) on snippet partitions gave a mean of $0.723$, ranging from $0.524$ to $0.957$ across annotator pairs (Figure~\ref{fig:iaa-reasoning-a}). The more constrained chunking decision therefore shows substantial agreement.

\paragraph{Pattern label:}
Pattern annotation was integrated into the same annotation pass under the written guidelines (Appendix~\ref{app:guidelines}). On a held-aside subset annotated by multiple annotators, pattern agreement was $\kappa \approx 0.59$, with the largest residual confusions between A (enumeration) and C (conclusion + premises) for snippets resembling both. The taxonomy and worked anchor cases were re-circulated for calibration before the production pass.

\subsection{Auto-Snippet Pipeline Quality}
\label{app:data-pipeline-quality}

We evaluate the \medsnip{} pipeline against human-annotated \medsnipbench{} across all 276 entries. \textit{Chunking fidelity} measures agreement with human snippet boundaries, while \textit{label robustness} measures sensitivity to the label-projection policy for auto-snippets.

\paragraph{Chunking fidelity:}
Against human snippet boundaries, \mbox{Mode 1} (atom-to-snippet) achieves a sentence-set F1 of $0.756$, compared with $0.734$ for \mbox{Mode 2} (snippet-direct). Mean embedding cosine similarity shows the same pattern at $0.824$ and $0.794$, respectively. 

\mbox{Mode 1} therefore more closely reproduces human snippet structure, consistent with its two-pass design that preserves atom-level boundaries as shown in Table~\ref{tab:pipeline-quality}.

\begin{table}[t]
\centering
\small
\begin{tabular}{lrrrr}
\toprule
\textbf{Mode} & \textbf{sent F1} & \textbf{sent P} & \textbf{sent R} & \textbf{embed cos} \\
\midrule
Mode 1 & $0.756$ & $0.723$ & $0.898$ & $0.824$ \\
Mode 2 & $0.734$ & $0.744$ & $0.793$ & $0.794$ \\
\bottomrule
\end{tabular}
\caption{Auto-snippet pipeline chunking fidelity against the human gold structure on all 276 entries.}
\label{tab:pipeline-quality}
\end{table}

\paragraph{Snippet count:}
The pipeline under-segments relative to humans.
Auto snippets total $1{,}578$ in \mbox{Mode 1} and $2{,}132$ in \mbox{Mode 2}, compared with $2{,}524$ human gold snippets, giving ratios of $0.62$ and $0.84$.
Consumer entries are reproduced well, with sentence F1 above $0.85$ in both modes, while vignettes are harder at $\approx 0.65$ in both modes.
This is consistent with the harder clinical-reasoning content noted in the subset-selection analysis above.

\paragraph{Label projection sensitivity.}
Auto-snippets inherit labels projected from their source atoms. We compare three policies. \textit{AND} requires all covered atoms to be true, with mixed-label sentences treated as false. \textit{Majority} uses the majority label, with ties resolved as true. \textit{Drop-mixed} removes auto-snippets spanning conflicting human labels and is used throughout the main body. AND and majority serve as robustness checks and agree in sign with drop-mixed in eleven of the twelve verifier-by-mode cells. Projection effects are small relative to model-level differences. Drop-mixed retains $1{,}382$ snippets in \mbox{Mode 1} and $1{,}753$ in \mbox{Mode 2}, with dropped cases reflecting mixed-label coverage rather than pipeline failures.

\section{Annotation Guidelines}
\label{app:guidelines}

The guidelines below were distributed to annotators with only operational and identifying details removed. The pattern taxonomy and worked anchor examples are reproduced unchanged.

\subsection{Background}
\label{app:gl-background}

The dataset contains 5{,}755 atomic claims from 276 medical QA entries. Annotators group these atoms into snippets and assign dual labels. Each entry shows:
\begin{itemize}[leftmargin=*,itemsep=2pt]
\item \textbf{Query}: the medical question.
\item \textbf{Full Response}: the LLM-generated answer.
\item \textbf{Atomic Claims}: statements extracted from the response with expert true or false labels.
\end{itemize}

\paragraph{Why we merge at all:}
Atomic-claim extraction can be too fine-grained. An atom such as \textit{Option B is the correct answer''} or \textit{The factors include the individual’s age’’} can be unverifiable without surrounding reasoning. Medical responses, especially clinical vignettes, often connect claims through \textit{because / since / therefore / requires}. Splitting these chains removes context needed for verification.

\begin{table*}[!t]
\centering
\footnotesize
\renewcommand{\arraystretch}{1.12}
\setlength{\tabcolsep}{3pt}

\begin{tabular}{@{}
p{0.05\textwidth}
p{0.08\textwidth}
p{0.61\textwidth}
p{0.20\textwidth}
@{}}
\toprule
\textbf{Code}
&
& \textbf{Example}
& \textbf{Decision and Rationale} \\
\midrule

\rowcolor{softgray}
\multicolumn{4}{c}{\textbf{Patterns that trigger MERGE}} \\
\midrule


\rowcolor{softgray!55}
\multicolumn{4}{c}{
\parbox{0.95\textwidth}{
\raggedright
\textcolor{darkgreen}{\textbf{Pattern A}}
\quad
\textbf{Enumeration of properties of one subject.}
Multiple features, factors, or properties describe the same subject.
The enumeration is the verifiable unit.
}
} \\
\midrule

\textcolor{darkgreen}{\textbf{A.1}}
&
&
\textit{Consumer query on safe Advil dose}

~

&
\multirow{6}{0.20\textwidth}{
\textcolor{darkgreen}{\textbf{MERGE.}}

Atoms 2--4 begin with ``The factors include\ldots'' and are meaningless without Atom 1.
}
\\

&
\textit{Atom 1}
&
``The safe amount of Advil (ibuprofen) to take at one time depends on several factors.''
&
\\

&
\textit{Atom 2}
&
``The factors include the individual's age.''
&
\\

&
\textit{Atom 3}
&
``The factors include the individual's weight.''
&
\\

&
\textit{Atom 4}
&
``The factors include the individual's overall health.''
&
\\

\cmidrule(lr){2-4}

\textcolor{darkgreen}{\textbf{A.2}}
&
&
\textit{Vignette on a 72-year-old man with suspected leukemia}

~

&
\multirow{5}{0.20\textwidth}{
\textcolor{darkgreen}{\textbf{MERGE.}}

Three clinical features that together form the ``CLL picture.''
}
\\

&
\textit{Atom 10}
&
``Chronic Lymphocytic Leukemia (CLL) typically presents with a high leukocyte count.''
&
\\

&
\textit{Atom 11}
&
``CLL typically presents with anemia.''
&
\\

&
\textit{Atom 12}
&
``CLL typically presents with thrombocytopenia.''
&
\\

\midrule


\rowcolor{softgray!55}
\multicolumn{4}{c}{
\parbox{0.95\textwidth}{
\raggedright
\textcolor{darkgreen}{\textbf{Pattern B}}
\quad
\textbf{Causal or conditional chain.}
Atoms are connected by \textit{because}, \textit{since}, \textit{as},
\textit{due to}, \textit{requires}, or \textit{therefore}, or one atom
serves as the premise for another.
}
} \\
\midrule

\textcolor{darkgreen}{\textbf{B.1}}
&
&
\textit{Vignette on a 23-year-old with treatment-resistant schizophrenia}

~

&
\multirow{6}{0.18\textwidth}{
\textcolor{darkgreen}{\textbf{MERGE.}}

Atom 7 states the condition; Atoms 8 and 9 specify the two required failures.
}
\\

&
\textit{Atom 7}
&
``Clozapine is generally reserved for patients who have failed to respond to at least two previous adequate trials of antipsychotic medications.''
&
\\

&
\textit{Atom 8}
&
``Patients must have failed at least one conventional antipsychotic.''
&
\\

&
\textit{Atom 9}
&
``Patients must have failed at least one second-generation antipsychotic.''
&
\\

\cmidrule(lr){2-4}

\textcolor{darkgreen}{\textbf{B.2}}
&
&
\textit{Consumer query on Hepatitis A IgM}

~

&
\multirow{4}{0.18\textwidth}{
\textcolor{darkgreen}{\textbf{MERGE.}}

Atom 8 is the reason for Atom 7.
}
\\

&
\textit{Atom 7}
&
``A non-reactive result does not necessarily rule out a past infection.''
&
\\

&
\textit{Atom 8}
&
``IgM antibodies may not be detectable after a certain period.''
&
\\

\midrule


\rowcolor{softgray!55}
\multicolumn{4}{c}{
\parbox{0.95\textwidth}{
\raggedright
\textcolor{darkgreen}{\textbf{Pattern C}}
\quad
\textbf{Conclusion + supporting premises.}
A recommendation, diagnosis, or final answer is meaningful only when
read with the reasoning or referent that supports it.
}
} \\
\midrule

\textcolor{darkgreen}{\textbf{C.1}}
&
&
\textit{Same schizophrenia vignette, end of response}

~

&
\multirow{3}{0.18\textwidth}{
\textcolor{darkgreen}{\textbf{MERGE.}}

Atom 20 is unverifiable without Atom 19.
}
\\

&
\textit{Atom 19}
&
``The most appropriate next step in management is to initiate clozapine.''
&
\\

&
\textit{Atom 20}
&
``Option B is the correct answer.''
&
\\

\cmidrule(lr){2-4}

\textcolor{darkgreen}{\textbf{C.2}}
&
&
\textit{Same vignette, earlier in response}

~

&
\multirow{6}{0.18\textwidth}{
\textcolor{darkgreen}{\textbf{MERGE.}}

Atoms 1--2 give the differential; Atom 3 is the conclusion drawn from it.
}
\\

&
\textit{Atom 1}
&
``The patient's aggressive behavior is likely secondary to schizophrenia.''
&
\\

&
\textit{Atom 2}
&
``The patient's aggressive behavior could be due to a psychotic disorder.''
&
\\

&
\textit{Atom 3}
&
``The most appropriate next step in management is to consider an alternative antipsychotic medication.''
&
\\

\bottomrule
\end{tabular}

\caption{\textbf{Structural patterns that trigger merging.}
Patterns A--C capture cases in which atomic decomposition removes context required for verification. The worked examples were used as inter-annotator anchors.}
\label{tab:annotation-merge-patterns}
\end{table*}

\begin{table*}[!t]
\centering
\footnotesize
\renewcommand{\arraystretch}{1.08}
\setlength{\tabcolsep}{3pt}

\begin{tabular}{@{}
p{0.05\textwidth}
p{0.08\textwidth}
p{0.61\textwidth}
p{0.20\textwidth}
@{}}
\toprule
\textbf{Code}
&
& \textbf{Example}
& \textbf{Decision and Rationale} \\
\midrule

\rowcolor{softgray}
\multicolumn{4}{c}{\textbf{Patterns that justify KEEPING ATOMIC}} \\
\midrule


\rowcolor{softgray!55}
\multicolumn{4}{c}{
\parbox{0.95\textwidth}{
\raggedright
\textcolor{darkblue}{\textbf{Pattern D}}
\quad
\textbf{Complete standalone fact, definition, or lab interpretation.}
The atom is a self-contained medical statement and requires no surrounding text for verification.
}
} \\
\midrule

\textcolor{darkblue}{\textbf{D.1}}
&
&
\textit{Hepatitis A IgM definition}

~

&
\multirow{4}{0.20\textwidth}{
\textcolor{darkblue}{\textbf{ATOMIC.}}

A complete definition; nothing else is needed.
}
\\

&
\textit{Atom 1}
&
``A non-reactive result for the Hepatitis A IgM (Immunoglobulin M) antibody test indicates that the individual does not have a recent or current Hepatitis A infection.''
&
\\

\cmidrule(lr){2-4}

\textcolor{darkblue}{\textbf{D.2}}
&
&
\textit{Lab value interpretation}

~

&
\multirow{4}{0.20\textwidth}{
\textcolor{darkblue}{\textbf{ATOMIC.}}

A single lab-value claim verifiable in isolation.
}
\\

&
\textit{Atom 1}
&
``A platelet count of 119{,}000/mm$^3$ is within normal limits.''
&
\\

&
&
\textit{This claim is false (normal range: 150{,}000--400{,}000), but factuality is a labeling question rather than a grouping decision.}
&
\\

\midrule


\rowcolor{softgray!55}
\multicolumn{4}{c}{
\parbox{0.95\textwidth}{
\raggedright
\textcolor{darkblue}{\textbf{Pattern E}}
\quad
\textbf{Topic genuinely shifts.}
The response moves from one subject to an unrelated one.
}
} \\
\midrule

\textcolor{darkblue}{\textbf{E}}
&
&
\textit{Example}
&
\multirow{5}{0.20\textwidth}{
\textcolor{darkblue}{\textbf{ATOMIC.}}

Split at the topic boundary.
}
\\

&
&
\begin{itemize}[leftmargin=*,itemsep=2pt, topsep=0pt]
    \item The response may shift from \textit{drug composition to overdose risk}, or from one differential diagnosis to a different one.
    \item The claims concern substantively different subjects and do not depend on one another for verification.
\end{itemize}
&
\\

\midrule


\rowcolor{softgray!55}
\multicolumn{4}{c}{
\parbox{0.95\textwidth}{
\raggedright
\textcolor{darkblue}{\textbf{Pattern F}}
\quad
\textbf{Same topic, but each atom carries a different checkable fact.}
Nearby atoms concern the same subject but each expresses an independently verifiable fact.
}
} \\
\midrule

\textcolor{darkblue}{\textbf{F}}
&
&
\textit{Example}
&
\multirow{7}{0.20\textwidth}{
\textcolor{darkblue}{\textbf{ATOMIC.}}

Keep independently verifiable facts separate.
}
\\

&
&
\begin{itemize}[leftmargin=*,itemsep=2pt, topsep=0pt]
    \item Two claims about the same drug can remain separate if they convey distinct information.
    \item A standalone fact does not need to be merged into a nearby enumeration merely because it appears next to one.
    \item The ``isolated [false] atom'' idiom also falls under Pattern F: a false atom is kept separate so that it does not contaminate adjacent true claims.
\end{itemize}

&
\\

\bottomrule
\end{tabular}

\caption{\textbf{Structural patterns that justify keeping atoms separate.}
Patterns D--F preserve claims that remain independently verifiable despite appearing near related content.}
\label{tab:annotation-atomic-patterns}
\end{table*}

We therefore merge atoms when separating them would break a logical dependency. Every rule below and each consensus judgment follows the same question:
\begin{quote}
\textit{``If I read only this atom (or only this snippet), do I have enough information to check whether it is medically correct?’’}
\end{quote}
If \textbf{yes}, keep it atomic. If \textbf{no}, merge it with the atoms that provide the missing context.

\subsection{Task}
\label{app:gl-task}

For each entry, the annotator defines shared context as key-value pairs describing patient demographics, medications, or topic, then groups related atoms into snippets using an editable LLM draft. Each snippet receives two labels, \emph{(i). with context}, indicating correctness for the patient or query, and \emph{(ii). in general}, indicating medical correctness. The annotator assigns a structural pattern (A–F), marks ambiguity when correctness cannot be determined, and adds notes when labels differ.

\subsection{Shared Context}
\label{app:gl-context}

Shared context is defined once per every entry and applies to all snippets, capturing information from the original query that may be needed to interpret and verify individual statements. For clinical vignettes, typical keys include \textit{age}, \textit{sex}, \textit{chief\_complaint}, \textit{medical\_history}, \textit{medications}, \textit{vitals}, \textit{treatment\_history}, and \textit{correct\_answer}. For consumer-health queries, where less patient-specific information is required, typical keys include \textit{topic}, \textit{drug\_a}, and \textit{drug\_b}.

\subsection{Grouping Rules}
\label{app:gl-grouping}

Tables~\ref{tab:annotation-merge-patterns} and~\ref{tab:annotation-atomic-patterns} give the six structural patterns used to operationalize this decision, together with the worked examples used as inter-annotator anchors. Default to atomic. It is cheaper and easier to merge during consensus review than to split a bad merge. If Pattern A, B, or C clearly fires, merge without hesitation. After the merge / keep-atomic decision, the annotator records the \textbf{dominant pattern code} (A--F) on the snippet. The code captures the structural reason for the chunking decision and enables per-pattern downstream analyses.

\paragraph{Hybrid snippets:}
A snippet may carry an optional secondary code when two patterns clearly apply. For example, a snippet describing a drug, its side effect, and the resulting need for monitoring is primarily a causal chain (B), while the monitoring atom also functions as a conclusion supported by the side-effect premise (C). 

This is recorded as primary B and secondary C. The secondary code is reserved for genuinely hybrid cases rather than routine close calls. When the choice is borderline, when a secondary code is set, or when none of the six codes fit cleanly, the annotator adds a pattern note (1--2 sentences) explaining the rationale. Genuine non-fits (truly outside A--F) are flagged for lead-researcher adjudication rather than forced into an ill-fitting code.

\subsection{Snippet Text}
\label{app:gl-snippet-text}

The snippet text should be a self-contained, verifiable statement that can be judged without reading the original query. An LLM-generated draft is provided for the annotator to review and edit. The annotator resolves pronouns using the shared context, replaces references such as the patient'' with a 23-year-old male,’’ keeps standalone facts unchanged, and does not introduce medical information absent from the source atoms.

\begin{figure*}[!t]
\centering
\includegraphics[width=0.95\linewidth]{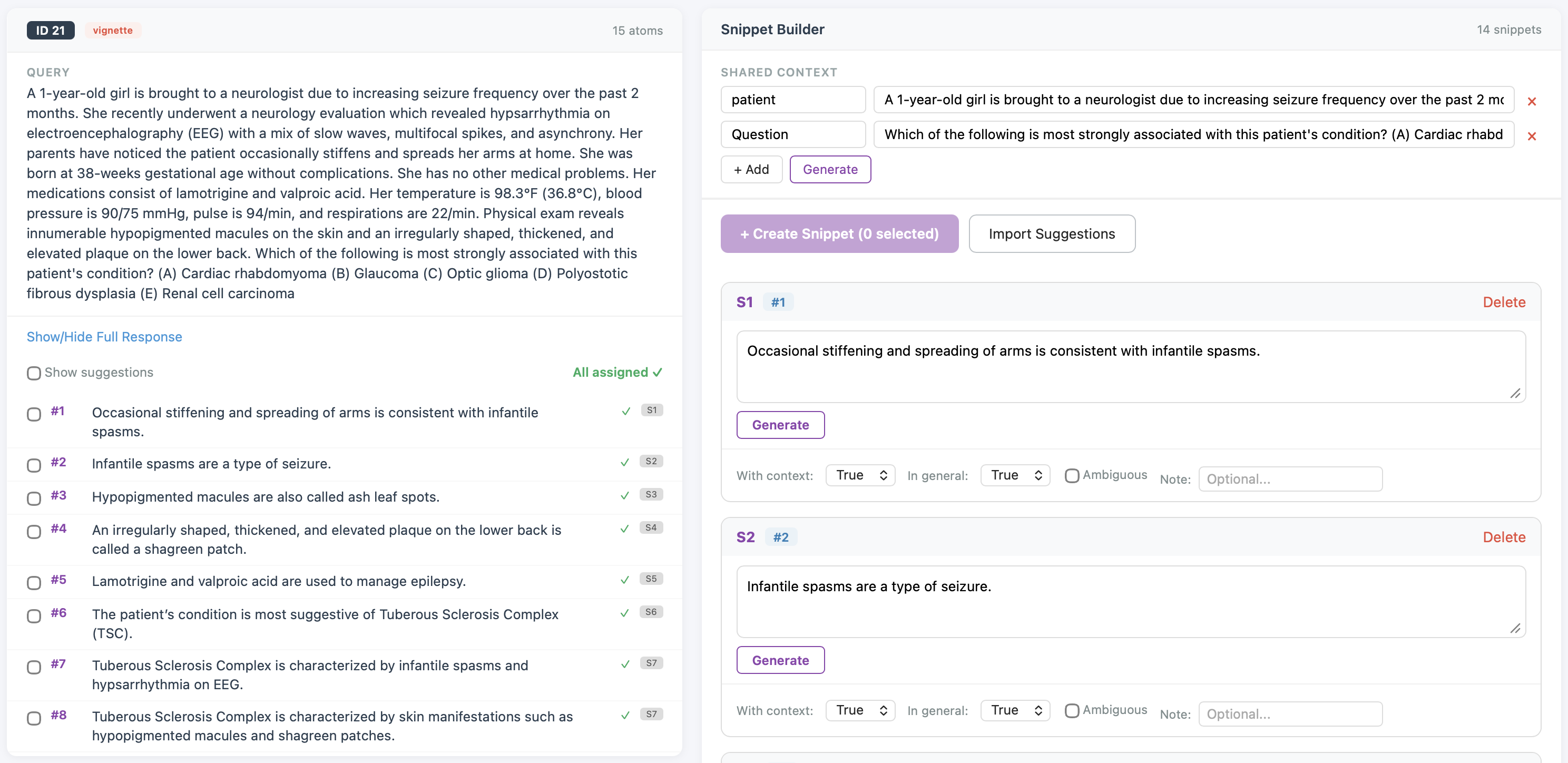}
\caption{\textbf{Annotation dashboard.}
The left panel shows atomic claims with expert labels, the query, and the full LLM response. The right panel shows shared context and per-snippet cards containing atom selection, merged text, dual labels, ambiguity status, notes, and pattern code. All annotator-identifying elements have been removed.}
\label{fig:dashboard}
\end{figure*}

\subsection{Dual Labels}
\label{app:gl-dual-labels}

Each snippet receives two labels. Here, context'' refers to information in the user query and earlier statements in the same response, excluding external medical knowledge and other entries. The \textit{in-general} label indicates whether the snippet is correct as a general medical statement. For example, Initiation of clozapine requires a 2-week trial’’ may be labeled \texttt{TRUE} in general if a short initial trial is medically reasonable. The \textit{with-context} label indicates whether the same snippet is correct for the specific patient or query. In a treatment-resistant schizophrenia vignette, the same statement may be labeled \texttt{FALSE} with context if the patient requires a longer trial. When the two labels differ, the statement is generally correct but contextually wrong, or vice versa, and the annotator adds a note explaining the difference. Labels are \textbf{True} for medically or factually correct statements, \textbf{False} for factual errors, and \textbf{Ambiguous} when correctness cannot be determined. Ambiguous labels also require a note.

\subsection{Annotation Dashboard}
\label{app:gl-dashboard}

Annotators use a browser-based dashboard implementing the workflow in Appendix \ref{app:gl-task}. As shown in Figure~\ref{fig:dashboard}, the interface presents the query, response, and atomic claims with expert labels, alongside controls for grouping atoms into snippets, editing snippet text, assigning dual labels and pattern codes, and recording ambiguity or notes. 

The dashboard ensures that all atoms are assigned before an entry is completed and exports annotations as JSON.

\section{Decomposition and Verification Prompts}
\label{app:prompts}

The prompts are built from reusable guideline, schema, formatting, granularity, and task-instruction components.
We describe each component once and provide a composition map linking production prompts to their components.

\subsection{Single-Call Baseline Verifier}
\label{app:prompts-baseline}

The baseline verifier (Section~\ref{subsec:verification-modes}) takes one unit and returns a binary verdict in a single LLM call. Atom and snippet baselines share these prompts; the unit text field is \texttt{claim} for atoms and \texttt{snippet\_text} for snippets. Figures~\ref{fig:prompt-claim-only} and \ref{fig:prompt-full-context} show both prompts verbatim.

\subsection{MedSNIP Pipeline: Prompt Composition}
\label{app:prompts-composition}

The pipeline (Section~\ref{subsec:medsnip-pipeline}) uses three system prompts: Mode~1 extract, Mode~1 cluster, and Mode~2 direct.
Each prompt is assembled from the reusable blocks in Table~\ref{tab:prompt-composition}.
Consumer and vignette variants use the same prompt skeleton, with subset-specific schemas, text rules, and granularity rules. The five reusable blocks are shown in Figures~\ref{fig:block-merge}--\ref{fig:block-extract-common}.

\begin{table}[!t]
\centering
\small
\renewcommand{\arraystretch}{1.08}
\setlength{\tabcolsep}{4pt}

\begin{tabularx}{\columnwidth}{@{}
>{\raggedright\arraybackslash}X
>{\centering\arraybackslash}p{0.18\columnwidth}
>{\centering\arraybackslash}p{0.18\columnwidth}
>{\centering\arraybackslash}p{0.18\columnwidth}
@{}}
\toprule

& \multicolumn{2}{c}{\textbf{Mode 1}}
& \textbf{Mode 2} \\
\cmidrule(lr){2-3}

\textbf{Component}
& \textbf{Extract}
& \textbf{Cluster}
& \textbf{Direct} \\
\midrule

Merge patterns
& -- & \checkmark & \checkmark \\

Atom extraction
& \checkmark & -- & -- \\

Granularity rules
& \checkmark & -- & -- \\

Context schema
& \checkmark & \checkmark & \checkmark \\

Snippet text rules
& -- & \checkmark & \checkmark \\

\midrule

\textbf{Output JSON}
& \texttt{extract}
& \texttt{cluster}
& \texttt{direct} \\

\bottomrule
\end{tabularx}

\caption{\textbf{Composition of reusable prompt components.}
Mode 1 separates atom extraction from clustering, whereas Mode 2 constructs snippets directly. Checkmarks indicate the components included at each stage.}
\label{tab:prompt-composition}
\end{table}

\begin{figure}[!h]
\centering
\begin{tcolorbox}[title={\claimonly{} baseline}, width=\linewidth, fonttitle=\bfseries]
\small
\textbf{System Prompt:}\\
You are a cautious medical fact-checking assistant. Given a claim, decide if the claim is true. Respond with exactly one word: \texttt{"true"} if the claim is fully supported, otherwise \texttt{"false"}. No other words.

\vspace{4pt}
\textbf{User Message:}
\begin{verbatim}
Claim:
{snippet_text}
\end{verbatim}
\end{tcolorbox}
\caption{Baseline verifier in \claimonly{} mode.}
\label{fig:prompt-claim-only}
\end{figure}

\begin{figure}[!h]
\centering
\begin{tcolorbox}[title={\fullcontext{} baseline}, width=\linewidth, fonttitle=\bfseries]
\small
\textbf{System Prompt:}\\
You are a cautious medical fact-checking assistant. You will be given a user question, a full answer text, and a single extracted claim (decomposed from the full answer text). Use the full answer text ONLY to decontextualize the claim. Then decide whether the interpreted claim is medically correct in the real world. Do NOT treat the full answer text as evidence the claim is true; it is context for interpretation. Respond with exactly one word: \texttt{"true"} or \texttt{"false"}. No other words.

\vspace{4pt}
\textbf{User Message:}
\begin{verbatim}
Question:
{query}

Full answer text (interpretation only):
{full_text}

Extracted claim:
{snippet_text}
\end{verbatim}
\end{tcolorbox}
\caption{Baseline verifier in \fullcontext{} mode. The full answer is for disambiguation only.}
\label{fig:prompt-full-context}
\end{figure}

\begin{figure}[!h]
\centering
\begin{tcolorbox}[title={Block: Merge Patterns A--F}, width=\linewidth, fonttitle=\bfseries]
\small\setlength{\parskip}{2pt}
You mirror a medical-text annotator. Group an LLM-generated medical response into \textbf{snippets}, coherent, self-contained, verifiable units. Guiding question for every claim: \textit{If this stood alone, could a reader verify it medically?} If no $\to$ merge.

~

Three patterns that trigger MERGE

\texttt{<Pattern A-C from Section 3>}

~

Three patterns that justify KEEPING ATOMIC

\texttt{<Pattern D-F from Section 3>}

~

\textit{Default.} When torn, keep atomic. 

~

Coverage: every unit appears in exactly one snippet.
\end{tcolorbox}
\caption{Shared annotation guidelines, used by Mode~1 cluster and Mode~2.}
\label{fig:block-merge}
\end{figure}

\begin{figure}[!h]
\centering
\begin{tcolorbox}[title={Block: Shared-context schemas}, width=\linewidth, fonttitle=\bfseries]
\small\setlength{\parskip}{2pt}
\textit{Consumer Keys:} topic, drug\_a, drug\_b, condition, symptom. Omit absent keys; values are short and concrete.

~

\textit{Vignette Keys:} age, sex, chief\_complaint, medical\_history, medications, vitals, exam\_findings, treatment\_history, correct\_answer. Omit absent keys.
\end{tcolorbox}
\caption{Per-subset \texttt{shared\_context} dict shapes. Plugged into every pipeline prompt.}
\label{fig:block-schema}
\end{figure}

\pagebreak

\begin{figure}[!h]
\centering
\begin{tcolorbox}[title={Block: Snippet text rules}, width=\linewidth, fonttitle=\bfseries]
\small\setlength{\parskip}{2pt}
\textit{Consumer.} Each snippet's \texttt{output} is a self-contained verifiable statement. A reader should judge correctness without the query. Do NOT introduce facts absent from the source.

~

\textit{Vignette.} Each snippet's \texttt{output} must be self-contained. Inline relevant patient demographics from \texttt{shared\_context} so a reviewer can judge it standalone (e.g., ``this patient'' $\to$ ``this 55-year-old male with right arm weakness''). Do NOT introduce facts absent from the source.
\end{tcolorbox}
\caption{Per-subset snippet wording rules. Used by Mode~1 cluster and Mode~2.}
\label{fig:block-text}
\end{figure}

\begin{figure}[!h]
\centering
\begin{tcolorbox}[title={Block: Granularity rules}, width=\linewidth, fonttitle=\bfseries]
\small\setlength{\parskip}{2pt}
\textit{Consumer (fine).} ``X can cause drowsiness, dizziness, and confusion'' $\to$ three atoms, one per side effect. ``Drug X is an antihistamine containing compound Z'' $\to$ one atom. Single-property sentences $\to$ one atom.

~

\textit{Vignette (coarse).} One atom per sentence by default. Causal chains and multi-property sentences stay as one atom. The final-answer sentence is its own atom. Each differential-option sentence becomes one atom.
\end{tcolorbox}
\caption{Per-subset atom granularity rules for consumer-health and clinical-vignette entries. Used by Mode1 step1 to guide atomic claim extraction.}
\label{fig:block-granularity}
\end{figure}

\begin{figure}[!h]
\centering
\begin{tcolorbox}[title={Block: Atom Extraction}, width=\linewidth, fonttitle=\bfseries]
\small\setlength{\parskip}{2pt}
You are an atomic-claim extractor. Given a user query and an LLM response pre-segmented into numbered sentences, produce a \texttt{shared\_context} dict plus a list of atomic claims that follow the sentence structure.

~

\textit{Atom definition.} Each atom states ONE medically-meaningful proposition, light-normalized from its sentence(s). Resolve pronouns. \textbf{Preserve the response's stance} --- do NOT correct claims that look wrong.

~

\textit{Sentence-atom mapping.} One sentence $\to$ one atom by default; one sentence $\to$ multiple atoms only for clear comma-separated enumerations; two consecutive sentences $\to$ one atom only for indivisible claims.

~

\textit{Coverage.} Every sentence index must appear in at least one atom's \texttt{source\_sentences}.
\end{tcolorbox}
\caption{Atom-definition and sentence-mapping rules for extracting self-contained atomic claims. Used by Mode~1 step~1 (extract).}
\label{fig:block-extract-common}
\end{figure}

\subsubsection*{Task Heads and Output Schemas}

Each prompt ends with a mode-specific task and JSON schema (Figures~\ref{fig:task-extract}--\ref{fig:task-direct}).

\begin{figure}[!h]
\centering
\begin{tcolorbox}[title={Task head: Mode~1 step~1 (extract)}, width=\linewidth, fonttitle=\bfseries]
\small\setlength{\parskip}{2pt}
\textit{Task.} Use the granularity rule for the subset; emit \texttt{shared\_context} per the schema; cover every sentence index.

\textit{Output (JSON only):}
\begin{verbatim}
{ 
    <JSON schema> 
}
\end{verbatim}
\end{tcolorbox}
\caption{Mode 1 step 1 task and output schema for atomic claim extraction. Combines with the extraction-common block, granularity rules, and the per-subset context schema.}
\label{fig:task-extract}
\end{figure}

\begin{figure}[!h]
\centering
\begin{tcolorbox}[title={Task head: Mode~1 step~2 (cluster)}, width=\linewidth, fonttitle=\bfseries]
\small\setlength{\parskip}{2pt}
Given the query, the pre-extracted \texttt{shared\_context}, and a numbered list of \texttt{atomic\_claims}, group atoms into snippets per Patterns A--F. Every atom index appears in exactly one snippet. Write each snippet's \texttt{output} per the text rules; tag the dominant pattern (A--F) and a short \texttt{notes} rationale.

\textit{Output (JSON only):}
\begin{verbatim}
{ <JSON schema> }
\end{verbatim}
\end{tcolorbox}
\caption{Mode 1 step 2 task and output. Combines with the merge-patterns block, the per-subset context schema, and snippet text rules.}
\label{fig:task-cluster}
\end{figure}

\begin{figure}[!h]
\centering
\begin{tcolorbox}[title={Task head: Mode~2 (snippet-direct)}, width=\linewidth, fonttitle=\bfseries]
\small\setlength{\parskip}{2pt}
Given the query and numbered sentences, in one pass: (1) extract \texttt{shared\_context}; (2) identify atomic claims and group them into snippets per Patterns A--F (every sentence index appears in at least one snippet); (3) write \texttt{output} per the text rules; (4) tag the dominant pattern.

\textit{Sentence-boundary default.} The default is one sentence $\to$ one snippet. Merge only when an A/B/C cue is explicit across sentences; if not, split. Topical similarity alone is NOT enough. A common failure mode is collapsing the whole response into one snippet --- do not do that. For $N$-sentence responses, the output usually has $0.6N$ to $N$ snippets.

\textit{Output (JSON only):}
\begin{verbatim}
{ 
    <JSON schema> 
}
\end{verbatim}
\end{tcolorbox}
\caption{Mode 2 task and output. Combines with the merge-patterns block, the per-subset context schema, and snippet text rules.}
\label{fig:task-direct}
\end{figure}

\subsection{Retrieval-Augmented Verifier}
\label{app:prompts-verifier}

The retrieval verifier (Section~\ref{subsec:verifier}) follows the prompt sequence in Figures~\ref{fig:verifier-framing}–\ref{fig:verifier-force} during iterative verification. If no verdict is reached, the force-final fallback in Figure~\ref{fig:verifier-force} is used.

\begin{figure}[!h]
\centering
\begin{tcolorbox}[title={Verifier: Answer, Abstain or Search}, width=\linewidth, fonttitle=\bfseries]
\small\setlength{\parskip}{2pt}
You are a rigorous medical fact-checker. You receive a snippet, the \texttt{shared\_context}, and any evidence gathered so far. Your job is to \textbf{catch wrong claims} --- false-negative detection is the critical metric.

\textit{Confidence:} 0.90--1.00 direct specific evidence; 0.70--0.89 strong with minor uncertainty; 0.50--0.69 leaning but not confident; $<$0.50 search more. Be calibrated, not optimistic.

\textit{Output (JSON only):}

\begin{verbatim}
{
  <JSON schema for Final Answer,
  Abstain and Search>
}
\end{verbatim}
\end{tcolorbox}
\caption{Verifier framing and the per-iteration JSON options with calibrated confidence bands.}
\label{fig:verifier-framing}
\end{figure}

\begin{figure}[!h]
\centering
\begin{tcolorbox}[title={Verifier: Must-Search, Refutation}, width=\linewidth, fonttitle=\bfseries]
\small\setlength{\parskip}{2pt}
\textit{Must-search triggers.} Do not short-circuit on parametric knowledge when the snippet has specific numbers (doses, ranges, percentages, durations, thresholds), equivalence claims between drugs/conditions/mechanisms, vignette-style clinical reasoning, recent guidelines, or any uncertainty on a named drug/dose/condition.

\textbf{Refutation-biased phrasing.} Look for evidence the snippet is \emph{wrong}, not for confirmation.

\textbf{Source.} \texttt{web} for consumer info, brand-name OTC, recent guidelines, recency-sensitive content. \texttt{pubmed} for primary clinical literature, RCT efficacy, mechanism, rare conditions; use keyword phrases, not full questions.
\end{tcolorbox}
\caption{Verifier search behavior: when to issue a search, how to phrase it, and which retriever to call.}
\label{fig:verifier-search}
\end{figure}

\begin{figure}[!h]
\centering
\begin{tcolorbox}[title={Verifier: General Truth Framing}, width=\linewidth, fonttitle=\bfseries]
\small\setlength{\parskip}{2pt}
\textit{General-truth framing.} Judge the snippet as a \emph{general} medical statement, NOT vignette applicability.

\textbf{Compound-claim rule.}
A multi-sub-claim snippet is true only if every \emph{material} sub-claim is true.
Material errors include the wrong drug class, effect direction, mechanism, or target population.
Ignore acceptable rounding, rare-exception generalizations, hedging, minor phrasing or mechanism imprecision, overlapping-category framing, and comparative edge cases when the core assertion holds.

\textbf{Mixed evidence.} If evidence neither clearly supports nor refutes a specific factual claim with named entities, numbers, or mechanism $\to$ \textbf{false}. Generic broadly-true statements with no specific content $\to$ \textbf{true} if nothing specific is wrong.
\end{tcolorbox}
\caption{Verifier truth-judgement rules for general medical statements, including compound-claim materiality and handling of mixed or inconclusive evidence.}
\label{fig:verifier-compound}
\end{figure}

\begin{figure}[!h]
\centering
\begin{tcolorbox}[title={Verifier: Force-final Fallback}, width=\linewidth, fonttitle=\bfseries]
\small\setlength{\parskip}{2pt}
You are a cautious medical fact-checker making a final decision; no more searches are allowed.

\textbf{Prefer abstain over guessed False.}
If the evidence does not directly refute a material claim, return \textbf{abstain}.
If the main assertion is defensible and only peripheral sub-claims are doubtful, return \textbf{true} under the compound-claim rule.

Only emit \texttt{final\_answer: false} with concrete refuting evidence, such as a wrong entity, out-of-range number, inverted mechanism, or misassigned category.
\end{tcolorbox}
\caption{Force-final fallback prompt used at the iteration cap when no verdict is reached. Biased toward abstention under residual uncertainty.}
\label{fig:verifier-force}
\end{figure}

\section{Experimental Details}
\label{app:experiments}

This appendix documents the experimental setup, statistical methodology, full per-cell numbers, and sensitivity analyses behind the main-body results.

\subsection{Compute and API}
\label{app:exp-compute}

\paragraph{Closed-source models:}
\gptfivefour{} (with reasoning effort high) and \gptfouro{} are accessed through the OpenAI API. \gptfivefour{} is the canonical decomposer across atom and snippet conditions to keep the comparison free of atomizer-quality confounds. Per-experiment dollar costs for both closed-source models are in Table~\ref{tab:full-cells}.

\paragraph{Open-weight models:}
\gemma{}, \gptoss{}, \llamaseventy{}, and \llamaeight{} are accessed through the HF Inference router with the OpenAI-compatible chat-completions endpoint, so the same prompt and parsing pipeline applies across all six models. End-to-end dollar costs per cell are reported in Table~\ref{tab:full-cells}.

\paragraph{Retrieval backends:}
The retrieval-augmented verifier uses Google Serper API (\$0.001 per query) and a public PubMed biomedical-literature API (free, abstracts only). Both return $k=3$ snippets per query. Each verified item issues between one and five retrieval rounds depending on confidence-gated stopping, for an average of $\approx 1.9$ retrievals per item.

\paragraph{Random seeds:}
All bootstrap analyses use a single fixed seed of 42. The train/dev/test partition was selected as the random-seed sweep winner of 202. Stochastic LLM calls use the API default sampling temperature.

\subsection{Bootstrap CI Methodology}
\label{app:exp-bca}

Every confidence interval for an \fF{} difference is a 95\% bootstrap interval over $B = 10{,}000$ replicates, with significance declared when the interval excludes zero.
All resampling is paired, so each replicate recomputes both arms on the same draw and $\Delta$ is a paired difference. Two interval types are used depending on the resampling unit.
For flat samples, as in the per-pattern comparisons of Table~\ref{tab:patterns}, we use bias-corrected and accelerated (BCa) intervals.
The bootstrap distribution of $\Delta$ can be skewed when either F1 approaches zero, as in small pattern cells or under weak verifiers, and BCa corrects for bias and skewness.

When replicates resample clusters, we instead use percentile intervals.
For Table~\ref{tab:f1-cross-model} and Table~\ref{tab:full-matrix}, we resample whole source entries on \medsnipbench{} and whole claims on the external corpora. This accounts for dependence among snippets from the same answer and for the different numbers of units produced by the two arms.
Because the resulting statistic does not have the flat structure required for the BCa jackknife correction, we report percentile intervals.
The paired clustering also tends to narrow the intervals, since errors shared across the two arms cancel within each draw. The bold cells of Table~\ref{tab:f1-cross-model} are the 17 of 42 whose interval excludes zero, 7 of 18 on \medsnipbench{}, 2 of 12 on \healthfc{}, and 8 of 12 on \medhallu{}.

\subsection{Full Per-Cell Results}
\label{app:exp-full-cells}

\begin{figure}[h]
\centering
\includegraphics[width=\columnwidth]{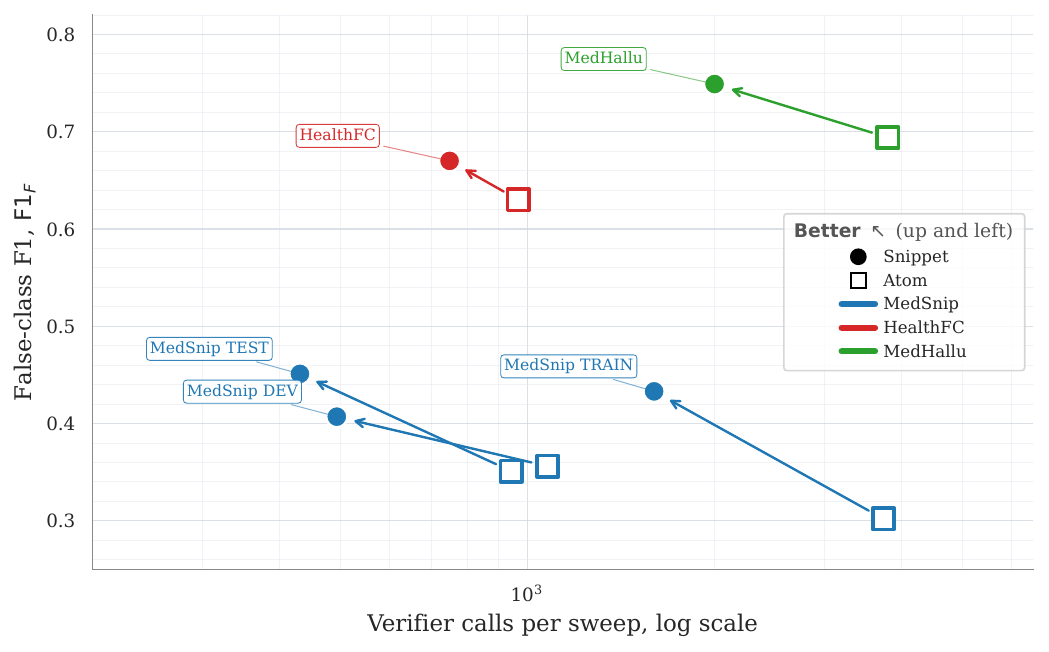}
\caption{Verifier calls versus \fF{} for snippet- and atom-level baselines across three datasets. Snippets require fewer calls while preserving or improving \fF{}.}
\label{fig:pareto-cost-f1}
\end{figure}

Table~\ref{tab:full-cells} provides the per-split \medsnipbench{} results underlying the aggregate numbers in Table~\ref{tab:f1-cross-model}.
It reports atom and snippet verifier-call counts, false-class F1, and dollar cost for each evaluated split.
Figure~\ref{fig:pareto-cost-f1} visualizes the corresponding cost--\fF{} trade-off, showing how snippet-level verification shifts the baselines toward fewer verifier calls

\begin{table*}[!t]
\centering
\small
\renewcommand{\arraystretch}{1.05}
\setlength{\tabcolsep}{4pt}

\begin{tabular}{@{}
l
l
r
r
r
r
r
r
r
@{}}
\toprule
\textbf{Model}
& \textbf{Split}
& \textbf{Atom calls}
& \textbf{Snippet calls}
& \textbf{Atom \fF{}}
& \textbf{Snippet \fF{}}
& \textbf{$\Delta$}
& \textbf{Atom cost}
& \textbf{Snippet cost} \\
\midrule


\rowcolor{softgray}
\multicolumn{9}{c}{
\textbf{\medsnipbench{} \claimonly{} Verification}
} \\
\midrule

\multirow{3}{*}{GPT-5.4 (high)}
& Train & 3,735 & 1,599 & 0.302 & 0.433
& \cellcolor{green!12}+0.131
& \$10.27 & \$7.13 \\
& Dev & 1,078 & 494 & 0.356 & 0.407
& \cellcolor{green!8}+0.051
& \$2.96 & \$2.20 \\
& Test & 942 & 431 & 0.351 & 0.451
& \cellcolor{green!12}+0.100
& \$2.59 & \$1.92 \\

\addlinespace[2pt]

\multirow{3}{*}{GPT-4o}
& Train & 3,735 & 1,599 & 0.328 & 0.416
& \cellcolor{green!10}+0.088
& \$1.70 & \$0.80 \\
& Dev & 1,078 & 494 & 0.361 & 0.395
& \cellcolor{green!8}+0.034
& \$0.49 & \$0.24 \\
& Test & 942 & 431 & 0.382 & 0.354
& \cellcolor{red!8}$-$0.028
& \$0.43 & \$0.22 \\

\addlinespace[2pt]

\multirow{3}{*}{\gemma{}}
& Train & 3,735 & 1,599 & 0.324 & 0.394
& \cellcolor{green!10}+0.070
& \$0.05 & \$0.02 \\
& Dev & 1,078 & 494 & 0.351 & 0.395
& \cellcolor{green!8}+0.045
& \$0.01 & $<\$0.01$ \\
& Test & 942 & 431 & 0.383 & 0.358
& \cellcolor{red!8}$-$0.025
& \$0.01 & $<\$0.01$ \\

\addlinespace[2pt]

\multirow{3}{*}{\gptoss{}}
& Train & 3,735 & 1,599 & 0.323 & 0.400
& \cellcolor{green!10}+0.077
& \$0.10 & \$0.06 \\
& Dev & 1,078 & 494 & 0.355 & 0.398
& \cellcolor{green!8}+0.043
& \$0.03 & \$0.02 \\
& Test & 942 & 431 & 0.406 & 0.408
& +0.002
& \$0.02 & \$0.02 \\

\addlinespace[2pt]

\multirow{3}{*}{\llamaseventy{}}
& Train & 3,735 & 1,599 & 0.299 & 0.322
& \cellcolor{green!8}+0.023
& \$0.05 & \$0.03 \\
& Dev & 1,078 & 494 & 0.326 & 0.369
& \cellcolor{green!8}+0.043
& \$0.02 & $<\$0.01$ \\
& Test & 942 & 431 & 0.368 & 0.314
& \cellcolor{red!8}$-$0.054
& \$0.01 & $<\$0.01$ \\

\addlinespace[2pt]

\multirow{3}{*}{\llamaeight{}}
& Train & 3,735 & 1,599 & 0.216 & 0.287
& \cellcolor{green!10}+0.071
& $<\$0.01$ & $<\$0.01$ \\
& Dev & 1,078 & 494 & 0.239 & 0.269
& \cellcolor{green!8}+0.029
& $<\$0.01$ & $<\$0.01$ \\
& Test & 942 & 431 & 0.316 & 0.255
& \cellcolor{red!10}$-$0.061
& $<\$0.01$ & $<\$0.01$ \\

\midrule


\rowcolor{softgray}
\multicolumn{9}{c}{
\textbf{\medsnipbench{} \fullcontext{} Verification}
} \\
\midrule

\multirow{3}{*}{GPT-5.4-high}
& Train & 3,735 & 1,599 & 0.475 & 0.475
& $-$0.000
& \$17.67 & \$11.27 \\
& Dev & 1,078 & 494 & 0.485 & 0.518
& \cellcolor{green!8}+0.034
& \$5.10 & \$3.48 \\
& Test & 942 & 431 & 0.491 & 0.491
& $-$0.000
& \$4.46 & \$3.04 \\

\addlinespace[2pt]

\multirow{3}{*}{GPT-4o}
& Train & 3,735 & 1,599 & 0.338 & 0.301
& \cellcolor{red!8}$-$0.037
& \$7.12 & \$3.25 \\
& Dev & 1,078 & 494 & 0.321 & 0.227
& \cellcolor{red!10}$-$0.094
& \$1.97 & \$0.96 \\
& Test & 942 & 431 & 0.222 & 0.135
& \cellcolor{red!10}$-$0.087
& \$1.79 & \$0.86 \\

\bottomrule
\end{tabular}

\caption{\textbf{Per-split verification results on \medsnipbench{}.}
Atom and snippet calls give verifier-call counts, with corresponding false-class \fF{}. $\Delta$ is the snippet–atom \fF{} difference, with gains and losses shown in green and red. Cost estimates one complete verification sweep.}

\label{tab:full-cells}
\end{table*}

\subsection{Aggregation Rule Sensitivity}
\label{app:exp-aggregation}

For datasets with answer-level labels, unit-level verdicts must be combined into a single prediction. We test whether this choice affects the snippet–atom comparison using five aggregation rules. \orfalse{} predicts \textsc{false} if any unit is false, following the \factscore{} default. \andfalse{} requires all units to be false, \majorityrule{} uses majority vote, \thresholdk{} predicts \textsc{false} when at least $k=2$ units are false, and a fifth rule takes the first unit alone.  

Across 14 dataset–granularity–mode–model settings, we recompute the snippet–atom \fF{} gap and its 95\% paired bootstrap interval under each rule.

\paragraph{Findings:}
The magnitude of the snippet–atom gap varies with the aggregation rule, but the advantage of snippets holds in most settings. Under \thresholdk{}, snippets outperform atoms in 12 of 14 settings with $P(\Delta > 0) \geq 0.95$. The two exceptions occur on \medsnipbench{} with \gptfouro{} in \fullcontext{} mode, where predictions are nearly degenerate. Under \orfalse{}, snippets win in 7 of 14 settings. The main exception is \andfalse{}, which reverses the comparison on \medhallu{} by $0.015$ \fF{} in favor of atoms ($P=0.036$). This rule favors atomization because it requires every unit to be \textsc{false}, and atomization produces more units.

\paragraph{Implication:}
Aggregation is therefore not a neutral implementation choice. It can change both the magnitude and direction of a granularity comparison, so decomposition-based evaluations should state the aggregation rule explicitly. The cost reduction reported in the main paper is unaffected because it depends only on the number of verification units.

\subsection{End-to-End Cost}
\label{app:exp-cost}

The cost analysis in Section~\ref{subsec:exp-f1} considers verification alone, but decomposition also adds overhead. An atom pipeline requires one decomposition call per answer, while atomize-then-group requires a second for grouping. Table~\ref{tab:cost-endtoend} includes this cost for \medsnipbench{} with \gptfivefour{} as the verifier. With \gptfivefour{}-high, decomposition costs \$23.63 and makes the full pipeline 19.1\% more expensive. With \gptoss{}, it costs only \$0.20, preserving nearly all of the 35.3\% verification saving. Snippet-direct further reduces this overhead by requiring only one decomposition call.

Call counts show the same pattern without relying on model prices. Atomize-then-group reduces total calls by 64.7\% on \medsnipbench{} but increases them by 14.5\% on \medhallu{} and 27.1\% on \healthfc{}. Snippet-direct instead yields reductions of 60.1\%, 20.9\%, and 17.3\%, respectively. Decomposition overhead therefore matters most for short answers, where fewer verification calls can be eliminated, consistent with Section~\ref{subsec:exp-structure}.

\begin{table}[!t]
\centering
\small
\renewcommand{\arraystretch}{1.08}
\setlength{\tabcolsep}{1.3pt}

\begin{tabularx}{\columnwidth}{
>{\raggedright\arraybackslash}X
>{\centering\arraybackslash}p{0.19\columnwidth}
>{\centering\arraybackslash}p{0.22\columnwidth}
>{\centering\arraybackslash}p{0.22\columnwidth}
}
\toprule
\textbf{Decomposer}
& \makecell{\textbf{Decomp.}\\\textbf{cost}}
& \makecell{\textbf{Verification}\\\textbf{reduction}}
& \makecell{\textbf{End-to-end}\\\textbf{reduction}} \\
\midrule

\gptfivefour{}-high
& \$23.63
& +54.3\%
& \cellcolor{red!10}$-$19.1\% \\

\gptfouro{}
& \$5.79
& +33.3\%
& \cellcolor{green!8}+8.8\% \\

\gemma
& \$0.29
& +22.6\%
& \cellcolor{green!12}+21.3\% \\

\llamaseventy{}
& \$0.21
& +26.0\%
& \cellcolor{green!12}+24.8\% \\

\gptoss{}
& \$0.20
& +35.3\%
& \cellcolor{green!15}+34.1\% \\

\llamaeight{}
& \$0.03
& +19.1\%
& \cellcolor{green!12}+18.9\% \\

\bottomrule
\end{tabularx}

\caption{End-to-end cost reduction from snippet-level verification across different decomposers, comparing verification cost savings before and after accounting for decomposition overhead.}
\label{tab:cost-endtoend}
\end{table}

\subsection{Wording Against Grouping}
\label{app:normalisation}

Snippet-level gains can come from grouping dependencies or making snippets easier to verify. Patterns D, E, and F provide a control because they pair one atom with one snippet, so gains must come from wording. To separate these effects, we rewrite each atom using decontextualization rules without grouping, then compare the raw atom, rewritten atom, and snippet.

\begin{table}[!t]
\centering
\small
\renewcommand{\arraystretch}{1.08}
\setlength{\tabcolsep}{1.5pt}

\begin{tabularx}{\columnwidth}{@{}
>{\raggedright\arraybackslash}X
>{\raggedright\arraybackslash}X
>{\centering\arraybackslash}X
>{\centering\arraybackslash}X
>{\centering\arraybackslash}X
>{\centering\arraybackslash}X
@{}}
\toprule
\textbf{Pattern}
& \textbf{Decision}
& \textbf{Raw}
& \textbf{Rewrite}
& \textbf{Snippet}
& \makecell{\textbf{Gain}} \\
\midrule

A & Merge & 0.267 & 0.292 & 0.387 & 21\% \\
B & Merge & 0.245 & 0.279 & 0.383 & 25\% \\
C & Merge & 0.302 & 0.354 & 0.426 & 42\% \\
D & Keep  & 0.250 & 0.336 & 0.364 & 75\% \\
F & Keep  & 0.543 & 0.624 & 0.621 & 103\% \\

\midrule

\textbf{A--C}
& \textbf{Merge}
& \textbf{0.273}
& \textbf{0.306}
& \textbf{0.399}
& \cellcolor{green!15}\textbf{26\%} \\

\textbf{D--F}
& \textbf{Keep}
& \textbf{0.365}
& \textbf{0.449}
& \textbf{0.454}
& \cellcolor{red!8}\textbf{94\%} \\

\bottomrule
\end{tabularx}

\caption{\textbf{Effect of decontextualization by structural pattern.}
Raw, rewritten, and snippet columns report \fF{}. Wording gain is the percentage of the raw-to-snippet improvement recovered by rewriting alone.}
\label{tab:normalisation}
\end{table}

Rewriting explains 94\% of the gain for keep-atomic patterns but only 26\% for merge patterns, leaving the remaining 74\% attributable to grouping related atoms. Pattern B shows the clearest effect, with wording explaining only 25\% of its gain, consistent with its causal and conditional structure where claims depend on one another. Pattern E has only 39 units and negative deltas throughout, so we do not interpret it further.

\begin{table*}[!t]
\centering
\small
\renewcommand{\arraystretch}{1.08}
\setlength{\tabcolsep}{5pt}


\begin{tabularx}{\textwidth}{
>{\raggedright\arraybackslash}p{0.17\textwidth}
*{6}{>{\centering\arraybackslash}X}
}
\toprule

\rowcolor{softgray}
\multicolumn{7}{c}{
\textbf{Mode 1: Atomize-then-Group}
} \\
\midrule

\textbf{Decomposer $\downarrow$}
& \multicolumn{6}{c}{\textbf{Verifier $\rightarrow$}} \\

&
\textbf{GPT-5.4}
& \textbf{GPT-4o}
& \textbf{\gemma{}}
& \textbf{\gptoss{}}
& \textbf{\llamaseventy{}}
& \textbf{\llamaeight{}} \\
\midrule

GPT-5.4-high
& \cellcolor{green!12}\textbf{+0.114}
& \cellcolor{red!8}$-$0.038
& \cellcolor{green!8}+0.014
& \cellcolor{red!8}$-$0.009
& \cellcolor{red!12}\textbf{$-$0.077}
& \cellcolor{red!8}$-$0.035 \\

GPT-4o
& \cellcolor{green!12}\textbf{+0.084}
& \cellcolor{red!8}$-$0.038
& \cellcolor{red!8}$-$0.005
& \cellcolor{green!8}+0.024
& \cellcolor{red!8}$-$0.045
& \cellcolor{green!8}+0.015 \\

\gemma{}
& \cellcolor{green!12}\textbf{+0.055}
& \cellcolor{red!8}$-$0.002
& \cellcolor{red!8}$-$0.025
& \cellcolor{red!8}$-$0.015
& \cellcolor{red!8}$-$0.044
& \cellcolor{red!8}$-$0.004 \\

\gptoss{}
& \cellcolor{green!12}\textbf{+0.066}
& \cellcolor{red!8}$-$0.036
& \cellcolor{red!8}$-$0.026
& \cellcolor{red!8}$-$0.001
& \cellcolor{red!12}\textbf{$-$0.068}
& \cellcolor{red!8}$-$0.044 \\

\llamaseventy{}
& \cellcolor{green!12}\textbf{+0.051}
& \cellcolor{red!8}$-$0.044
& \cellcolor{red!8}$-$0.021
& \cellcolor{red!8}$-$0.013
& \cellcolor{red!12}\textbf{$-$0.063}
& \cellcolor{red!8}$-$0.032 \\

\llamaeight{}
& \cellcolor{green!8}+0.032
& \cellcolor{red!8}$-$0.008
& \cellcolor{green!8}+0.017
& \cellcolor{red!8}$-$0.018
& \cellcolor{red!8}$-$0.004
& \cellcolor{green!8}+0.036 \\

\bottomrule
\end{tabularx}

\vspace{7pt}


\begin{tabularx}{\textwidth}{
>{\raggedright\arraybackslash}p{0.17\textwidth}
*{6}{>{\centering\arraybackslash}X}
}
\toprule

\rowcolor{softgray}
\multicolumn{7}{c}{
\textbf{Mode 2: Snippet-Direct}
} \\
\midrule

\textbf{Decomposer $\downarrow$}
& \multicolumn{6}{c}{\textbf{Verifier $\rightarrow$}} \\

&
\textbf{GPT-5.4}
& \textbf{GPT-4o}
& \textbf{\gemma{}}
& \textbf{\gptoss{}}
& \textbf{\llamaseventy{}}
& \textbf{\llamaeight{}} \\
\midrule

GPT-5.4-high
& \cellcolor{green!12}\textbf{+0.099}
& \cellcolor{red!8}$-$0.027
& \cellcolor{green!8}+0.038
& \cellcolor{green!8}+0.022
& \cellcolor{red!8}$-$0.037
& \cellcolor{red!8}$-$0.025 \\

GPT-4o
& \cellcolor{green!12}\textbf{+0.092}
& \cellcolor{red!8}$-$0.032
& \cellcolor{red!8}$-$0.005
& \cellcolor{green!8}+0.033
& \cellcolor{red!8}$-$0.029
& \cellcolor{green!8}+0.031 \\

\gemma{}
& \cellcolor{green!12}\textbf{+0.061}
& \cellcolor{green!8}+0.005
& $-$0.000
& \cellcolor{red!8}$-$0.002
& \cellcolor{red!8}$-$0.023
& \cellcolor{green!8}+0.035 \\

\gptoss{}
& \cellcolor{green!12}\textbf{+0.055}
& \cellcolor{green!8}+0.005
& \cellcolor{green!8}+0.002
& \cellcolor{red!8}$-$0.025
& \cellcolor{red!8}$-$0.033
& \cellcolor{green!8}+0.030 \\

\llamaseventy{}
& \cellcolor{green!12}\textbf{+0.095}
& \cellcolor{red!8}$-$0.050
& \cellcolor{green!8}+0.001
& \cellcolor{green!8}+0.001
& \cellcolor{red!12}\textbf{$-$0.059}
& \cellcolor{green!8}+0.037 \\

\llamaeight{}
& \cellcolor{green!12}\textbf{+0.045}
& \cellcolor{red!8}$-$0.018
& \cellcolor{green!8}+0.012
& \cellcolor{green!8}+0.025
& \cellcolor{red!8}$-$0.018
& \cellcolor{green!8}+0.011 \\

\bottomrule
\end{tabularx}

\caption{\textbf{Snippet--atom \fF{} gap across decomposer--verifier pairs on \medsnipbench{}.}
Rows denote decomposers and columns denote verifiers. Green cells indicate gains and red cells indicate losses. Bold values have 95\% entry-clustered bootstrap confidence intervals excluding zero, based on 10{,}000 resamples.}
\label{tab:full-matrix}
\end{table*}

There are two caveats. For 382 of 4{,}868 merge atoms, the rewriter used information from the shared context to effectively reconstruct the merge. We exclude these cases because they do not provide a valid wording-only ablation. In addition, repeating verification on identical text produces a baseline difference of $0.013$ \fF{}, so smaller differences are not informative.

\subsection{Full Decomposer-by-Verifier Matrix}
\label{app:matrix}

Section~\ref{subsec:exp-robustness} reports a subset of the full decomposer-by-verifier experiment. Table~\ref{tab:full-matrix} shows all six decomposers paired with all six verifiers, under both pipeline modes on \medsnipbench{}. Each cell reports the difference between snippet and expert-atom \fF{} for the same verifier. Because \medsnipbench{} inherits its atomic claims from \citet{kim2025rethinking}, all decomposers are compared against the same human-annotated atom baseline. The results show that the verifier matters more than the decomposer. All significant gains occur with \gptfivefour{}, while all significant losses occur with \llamaseventy{}. No other verifier produces a significant difference. With \gptfivefour{}, the six decomposers yield gains from $+0.032$ to $+0.114$, with all but one reaching significance. Even the 31B and 8B open-weight decomposers therefore produce snippets that outperform expert atoms. 

Across verifiers, however, this advantage disappears, suggesting that gains depend mainly on the verifier’s ability to use the recovered structure rather than the model used to recover it. Mode 1 and Mode 2 perform similarly overall, supporting our decision to report both rather than select a default.

\section{Retrieval-Augmented Verifier Development Checks}
\label{app:verifier}

The main paper reports the final retrieval-augmented verifier and its coverage–\fF{} trade-off.
During development, we also tested two variants intended to recover full coverage while preserving the benefits of confidence gating.
The first used the retrieval verifier when its confidence exceeded the threshold and otherwise fell back to the single-call baseline.
The second routed examples by source subset, using the retrieval verifier for one subset and the single-call baseline for the other.
Both variants improved performance on the development split but did not provide a stable or consistent improvement on the held-out test split.
We therefore report the simpler confidence-thresholded verifier in the main paper and treat these full-coverage routing and ensembling strategies as future work rather than part of the core result.

\end{document}